\documentclass{article}

\PassOptionsToPackage{numbers, compress}{natbib}

\usepackage[preprint]{neurips_2024}

\usepackage[utf8]{inputenc} 
\usepackage[T1]{fontenc}    
\usepackage{hyperref}       
\usepackage{url}            
\usepackage{booktabs}       
\usepackage{amsfonts}       
\usepackage{nicefrac}       
\usepackage{microtype}      
\usepackage{xcolor}         
\usepackage{graphicx}
\usepackage{url}   

 \usepackage{amsmath}
  \usepackage{multirow} 
  \usepackage{mathtools,tabularx}
\usepackage{wrapfig}
 \newcommand\norm[1]{\left\lVert#1\right\rVert}
\newcommand\ourmethod{\textsc{TracksTo4D}}
\usepackage{longtable}
\newcommand{\vvv}[1]{\mathbf{#1}}

\newcommand{\linkk}[1]{\emph{\textcolor{magenta}{#1}}}
\title{Fast Encoder-Based 3D from Casual Videos via Point Track Processing}

\author{%
  \hskip 0.6cm Yoni Kasten$^1$  \\
  \And
  \hskip 1cm Wuyue Lu$^2$ \\
  \And
  \hskip 1cm Haggai Maron$^{1,3}$ 
  \AND
  \vspace{-25pt} \\
  $^1$NVIDIA Research \qquad
  $^2$Simon Fraser University \qquad
  $^3$Technion
}

\begin{document}

\maketitle

\begin{abstract}
 This paper addresses the long-standing challenge of reconstructing 3D structures from videos with dynamic content. Current approaches to this problem were not designed to operate on casual videos recorded by standard cameras or require a long optimization time. 
  Aiming to significantly improve the efficiency of previous approaches, we present \ourmethod, a learning-based approach that enables inferring 3D structure and camera positions from dynamic content originating from casual videos using a single efficient feed-forward pass. To achieve this, we propose operating directly over 2D point tracks as input and designing an architecture tailored for processing 2D point tracks. Our proposed architecture is designed with two key principles in mind: (1) it takes into account the inherent symmetries present in the input point tracks data, and (2) it assumes that the movement patterns can be effectively represented using a low-rank approximation.\ourmethod\ is trained in an unsupervised way on a dataset of casual videos utilizing only the 2D point tracks extracted from the videos, without any 3D supervision. Our experiments show that \ourmethod\ can reconstruct a temporal point cloud and camera positions of the underlying video with accuracy comparable to state-of-the-art methods, while drastically reducing runtime by up to 95\%.  We further show that \ourmethod\  generalizes well to unseen videos of unseen semantic categories at inference time. Project page: \href{https://tracks-to-4d.github.io/}{ \linkk{https://tracks-to-4d.github.io}.}  \vspace{-1mm}
  \end{abstract}

\section{Introduction}
\label{sec:intro}
Predicting 3D geometry in dynamic scenes is a challenging problem. In this problem setup, we are given access to multiple images of a scene taken sequentially, e.g., from a monocular video camera, where \emph{both} the content in the scene and the camera are moving. Our task is to reconstruct the dynamic 3D positions of the points seen in the images and the camera poses. This fundamental problem has gained significant interest from the research community over the years \cite{bregler2000recovering,parashar2017isometric,kumar2022organic,zhang2022structure}, mainly due to its important applications in many fields such as robot navigation, autonomous driving and 3D reconstruction of general environments \cite{jensen2021benchmark}. Importantly, in contrast to static scenes where the epipolar geometry constraints hold between the corresponding points of different views \cite{hartley2003multiple}, determining the depth of a moving point from monocular views is an ill-posed problem \cite{akhter2008nonrigid}. This causes standard Structure from Motion techniques \cite{schonberger2016structure, wu2013towards, mur2015orb} to be inadequate in this setup \cite{kopf2021robust}. 

\paragraph{Previous work and limitations.}  Many existing approaches for the above problem make simplifying assumptions that limit their applicability to real-world scenarios. Methods based on orthographic camera models and low-rank assumptions use matrix factorization techniques \cite{bregler2000recovering,kumar2022organic}, but the orthographic camera assumption might not be realistic and may cause reconstruction errors. Techniques that incorporate depth priors often require lengthy optimization processes in order to make the depth estimates across frames consistent \cite{kopf2021robust,zhang2022structure}. Other physics-based approaches make assumptions about rigid bones \cite{yang2022banmo,yang2021lasr} or isometric deformable surfaces \cite{parashar2017isometric} and typically involve complex, slow optimization per video. In addition, they may require foreground-background segmentation of the moving content, which is not always easily obtained. Alternatively, some methods are specifically tailored to certain object classes like humans \cite{wan2021encoder}, restricting their domain to those limited cases. Consequently, these prior methods are either not directly applicable to casual videos, or require long optimization time per video.

\begin{figure}[t!]
    \centering
  \includegraphics[width=0.99\textwidth]{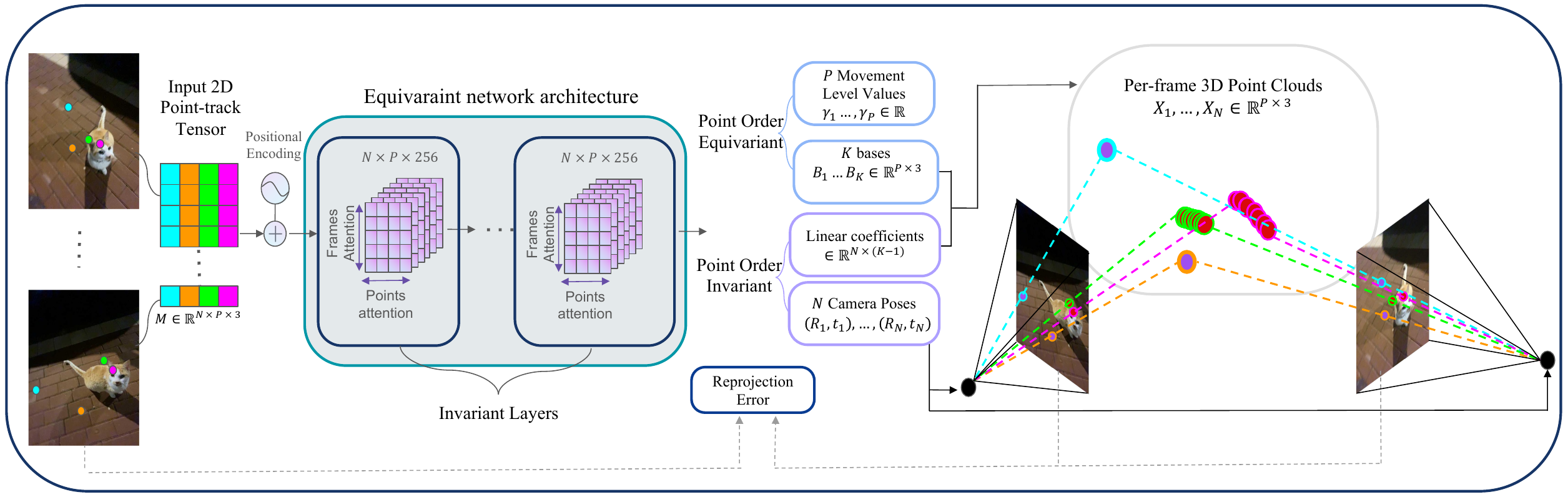}
  \caption{\small We present \ourmethod, a method for mapping a set of 2D point tracks extracted from casual dynamic videos into their corresponding 3D locations and camera motion. At inference time, our network predicts the dynamic structure and camera motion in a single feed-forward pass. Our network takes as input a set of 2D point tracks (left) and uses several multi-head attention layers while alternating between the time dimension and the track dimension (middle). The network predicts cameras, per-frame 3D points, and per-world point movement value (right). The 3D point internal colors illustrate the predicted 3D movement level values, such that points with high/low 3D motion are presented in red/purple colors respectively. These outputs are used to reproject the predicted points into the frames for calculating the reprojection error losses. See details in the text. The reader is encouraged to watch the supplementary video visualizations.\vspace{-5mm}}
  \label{fig:teaser}
\end{figure}

\paragraph{Our approach.} We propose \ourmethod,\footnote{4D since we have three Euclidean coordinates with an additional time coordinate} a novel approach for fast reconstruction of sparse dynamic 3D point clouds and camera poses from casual videos. Our main idea is to train a neural network on multiple videos to learn the mapping from the input image sequence to a sequence of the scene's 3D point clouds and camera poses. After training, the trained network can be efficiently applied to new image sequences using a single feed-forward pass, avoiding costly optimization. 

To enhance the method's ability to generalize across different types of videos and scenes, we made a crucial design choice: our approach processes point track tensors as input, rather than operating directly on the image sequence. Specifically, Each entry $(n,p)$ in these tensors represents the 2D position of a tracked point $p$ in a specific video frame $n$ \cite{bregler2000recovering}. Our main insight is that point track tensors may exhibit more common motion patterns across casual video domains compared to image pixels. In other words, we argue that processing the raw point track data rather than scene-specific pixels or features may enable learning class and scene-agnostic internal feature representations for improved generalization.
Importantly, recent advances in point tracking \cite{doersch2023tapir,karaev2023cotracker} enable efficiently inferring these point tracks from casual videos using pre-trained models. These two properties make point track matrices an attractive input for our learning method. 

Following this design choice, we design our architecture according to two principles:  (1) process point track tensors, which have a unique structure, and (2) encode meaningful prior knowledge about the reconstruction problem, as the problem is ill-posed in general.  In the following, we address these desired properties. 




First, we design a network architecture that can effectively and efficiently handle point track inputs. To do that, we propose a novel layer design that takes into account the symmetries of the problem: the mapping we aim to learn, from point track matrices to 3D point clouds and camera poses, preserves two natural symmetries: (i) the points being tracked can be arbitrarily permuted without affecting the problem; (ii) the frames containing these points exhibit temporal structure, adhering to an approximate time-translation symmetry. Following the Geometric Deep Learning paradigm \cite{bronstein2021geometric}, we build upon recent theoretical advances in equivariant learning \cite{maron2020learning} and integrate these two symmetries into our network architecture using dedicated attention and positional encoding mechanisms.

Second, a key challenge in predicting 3D dynamic motion and camera poses from 2D point tracks is that this problem is inherently ill-posed without additional constraints \cite{akhter2008nonrigid}. To address this, we integrate a low-rank movement assumption into our architecture, following the seminal work of \cite{bregler2000recovering}  which constrained output point clouds to be linear combinations of basis elements. Specifically, given an input point track tensor, our architecture equivariantly predicts a small set of input-specific basis elements. The output point clouds at each time frame are then defined as a linear combination of these basis elements, with the coefficients also predicted by the network. Notably, the first basis is assumed to fully represent the 3D static points in the video, while the remaining basis elements capture the 3D dynamic deviations. 
This structure effectively restricts the predicted point clouds to have a more specific form, making the problem more constrained.


Our approach is trained on a dataset of extracted point track matrices \cite{karaev2023cotracker} from raw videos without any 3D supervision by simply minimizing the reprojection errors, aiming to predict output point clouds that, after undergoing a perspective projection, will return the original 2D point tracks.
In our experiments, \ourmethod\ is trained on the Common Pets dataset \cite{sinha2023common}. 
We evaluate our method on test data with GT camera poses and GT depth information for point tracks, and demonstrate that it produces comparable results to state-of-the-art methods, while having a significantly shorter inference time by up to $95\%$. In addition, we show the method's ability to generalize to out-of-domain videos. 

\paragraph{Contributions.} In summary, our contributions are (1) A novel modeling of the dynamic reconstruction problem via learning on point tracks without 3D supervision; (2) A novel deep learning architecture incorporating two key principles: accounting for the symmetry of the data and encoding low-rank structure in the predicted point clouds  (3) Experiments demonstrating extremely fast inference time compared to baselines, accurate results, and strong generalization across other categories. 
\vspace{-1.5mm}
\section{Method}
\vspace{-1.5mm}
\label{sec:method}
         

\paragraph{Problem formulation.} Given a video of $N$ frames, let $M\in \mathbb{R}^{N\times P \times 3 } $ be a pre-extracted $2D$ point tracks tensor (Fig.~\ref{fig:teaser}, left side). This tensor represents the two-dimensional information about a set of $P$  world points that are tracked throughout the video. Each element in the tensor, $M_{i,j,:}$, stores three values: $(x,y,o)$ where $x,y\in \mathbb{R}$ are respectively the observed horizontal and vertical locations of point $j$ in frame $i$, and $o\in \{0,1\}$ indicates whether point $j$ is observed in frame $i$ or not. Our goal is to train a deep neural network to map the input point tracks tensor $M$ into a set of per-frame camera poses  $ \{R_i(M), \vvv{t}_i(M)\}_{i=1}^N$ and per-frame 3D points $\{X_i(M)\}_{i=1}^N$, where $R_i(M) \in \mathbb{SO}(3) , \vvv{t}_i(M)\in \mathbb{R}^3, X_i(M)\in \mathbb{R}^{P\times 3}$  (Fig.~\ref{fig:teaser}, right side).

\paragraph{Overview of our approach.}Our method receives $M\in \mathbb{R}^{N \times P \times 3}$ as input. This tensor is being processed by a neural architecture composed of multi-head attention layers where the attention is applied in an alternating fashion on the $P$ and the $N$ dimensions in each layer. These layers are defined in Sec.~\ref{section::architecture}. After a composition of several such layers, the network uses the resulting features in $\mathbb{R}^{N \times P \times d}$ to predict $N$ camera poses in $\mathbb{SO}^3\times \mathbb{R}^3$ and $N$ point clouds in $\mathbb{R}^{N\times P \times 3 }$. These $N$ point clouds are parameterized as a linear combination of  $K$ input specific point cloud bases $B_1(M),\dots B_K(M) \in \mathbb{R}^{P\times 3}$. This is discussed in detail in Sec.~\ref{section::output_parameterization}.  Our network is trained in an unsupervised way on a dataset of videos by minimizing the reprojection error and other regularization losses (Sec.~\ref{section::training}) that are used to update the model parameters. Our pipeline is illustrated in Fig.~\ref{fig:teaser}
\vspace{-1.5mm}
\subsection{Equivariant layers for point track tensors}
\vspace{-1.5mm}
\label{section::architecture}
 Following the geometric deep learning paradigm, our goal is to design a neural architecture that respects the underlying symmetries and structure of the data.
 
\begin{wrapfigure}[13]{R}{5.5cm}
    \centering
    \vspace{-20pt}
    \includegraphics[width=0.4\textwidth]{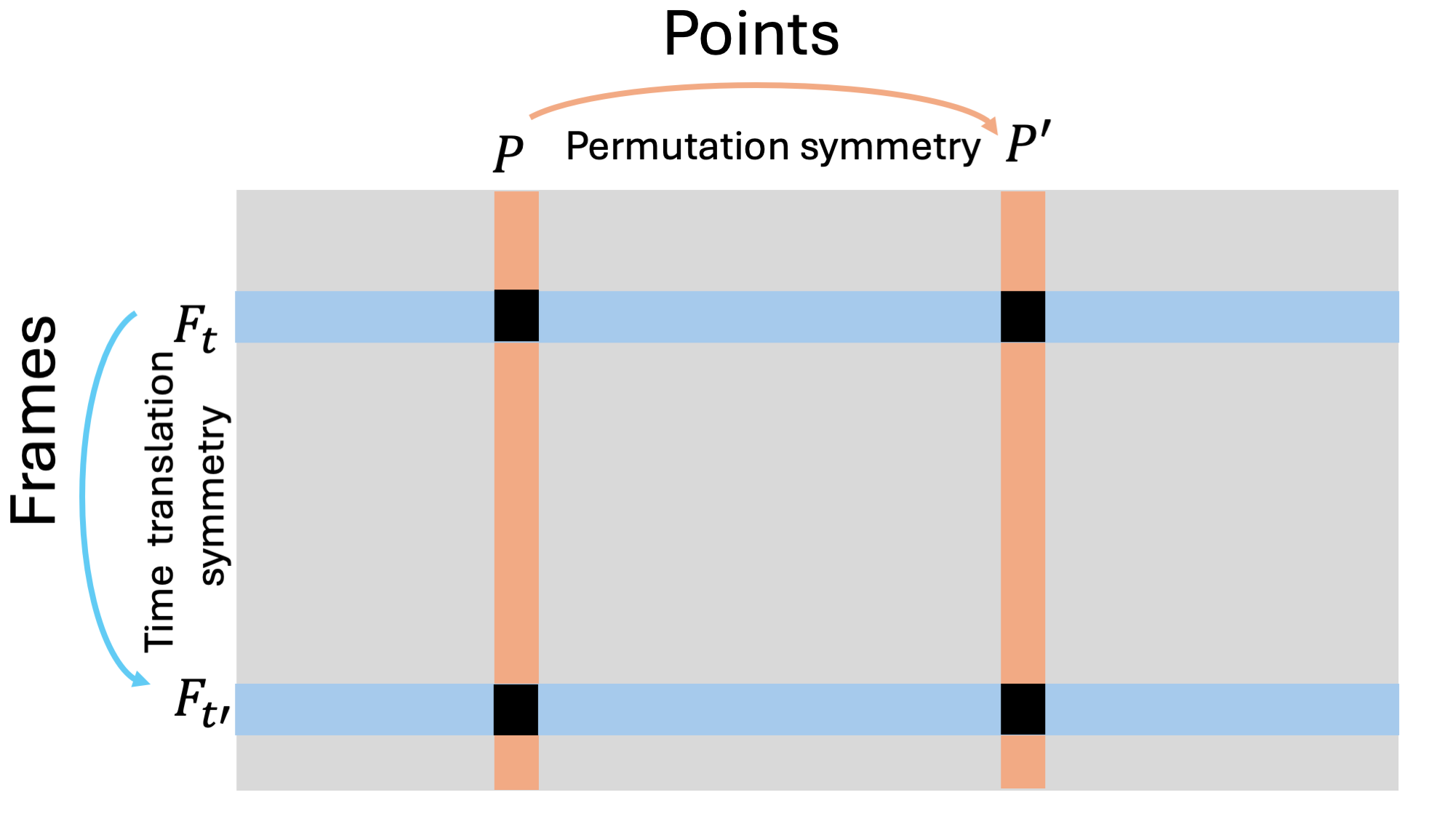}
    \caption{The symmetry structure of our problem. Frames (vertical) have time translation symmetry while points (horizontal) have set permutation symmetry.\vspace{-10mm}}
    \label{fig:sym}
\end{wrapfigure}

\paragraph{Symmetry analysis.} Our input is a tensor $M \in \mathbb{R}^{N\times P \times 3}$ representing a sequence of $N$ frames each with $P$ point coordinates. This structure gives rise to two key symmetries:
First, the order of the $P$ points within each frame does not matter - in other words,  permuting this axis results in an equivalent problem \cite{maron2020learning}. Formally, this axis has a permutation symmetry $S_P$ where $S_P$ is the symmetric group on $P$ elements.
 Second, along the temporal $N$ axis, we have an approximate translation symmetry arising from the ordered video sequence. This means that shifting the time frames is required to result in the same shift in our output. We model this with a cyclic group $C_N$ of order $N$. Both symmetries are illustrated in Fig.~\ref{fig:sym}. We note that while the cyclic group assumption may not be entirely accurate, we still find it useful as it helps us to derive appropriate parametric layers for our data, similar to how the convolutional layer is derived for data with translational symmetries such as images.
Taken together, the full symmetry group of the input space is the direct product $\mathcal{G}=C_N \times S_P$ combining these time and point permutation symmetries, acting on $\mathbb{R}^{N\times P \times 3}$ by $((t,\sigma)\cdot M)_{n,p,j}=M_{t^{-1}(n),\sigma^{-1}(p),j}$ for $(t,\sigma)\in \mathcal{G}$\footnote{This is different from the symmetry group studied in \cite{moran2021deep}, where the temporal structure was not exploited.}. 
Next, we will design an architecture equivariant to $\mathcal{G}$, to ensure that the model takes into account the symmetries above. 

\paragraph{Linear equivariant layers.} Point track tensors can be viewed as a collection of $N$ individual point tracks, each of which exhibits translational symmetry. The scenario where an object comprises a set of elements with their own symmetry group, such as a set of images or graphs, was previously explored in \cite{maron2020learning}. In that work, the authors characterized the general linear equivariant layer structure in such cases, termed the Deep Sets for Symmetric Elements (DSS) layer.
Building on the DSS approach, our basic linear equivariant layer for the point track tensors $M$ would take the form:
\begin{equation}\label{eq:DSS}
F(M)_{:,j} = L_1(M_{:,j}) + \sum_{j'=1}^{P} L_2(M_{:,j'})
\end{equation}

 \noindent where $L_i$ are linear translation equivariant function (i.e. convolutions), $M_{:,j} \in \mathbb{R}^{N\times d}$ are the columns of $M$ representing all the inputs for a specific tracked point,  $F(M)_{:,j}\in \mathbb{R}^{N\times d'}$ is the output column and $d,d'$ are the input and output feature channels respectively. To construct a neural network,  these layers can be interleaved with pointwise nonlinearities, similar to basic convolutional neural networks.
 
 \paragraph{Implementation via transformers and positional encoding.} While the linear layer design is reasonable, it may not be the optimal choice. To enhance the model, we design a new layer whose structure follows Equation \eqref{eq:DSS}, but incorporates nonlinear layers in the form of transformers \cite{vaswani2017attention}. Specifically, our layer $F$ is formulated similarly to Equation \eqref{eq:DSS}, but instead of convolutions ($L_i$) and summations, it utilizes two self-attention mechanisms and suitable temporal positional encoding across the $N$ dimension. Formally, our basic layer $F: \mathbb{R}^{N \times P \times d} \rightarrow \mathbb{R}^{N \times P \times d'}$ is  computed via four steps, which are described below:
\begin{equation}\label{eq:1}
     \vvv{\bar{q}}_{ij} = \bar{W}^QM_{ij},~\vvv{\bar{k}}_{ij} = \bar{W}^KM_{ij},~  \vvv{\bar{v}}_{ij} = \bar{W}^VM_{ij}
\Rightarrow
    \bar{M}_{ij}=  \sum_{i'=1}^N \frac{\exp(\vvv{\bar{q}}_{ij} \cdot \vvv{\bar{k}}_{i'j})}{\sum_{l=1}^N \exp(\vvv{\bar{q}}_{ij} \cdot \vvv{\bar{k}}_{lj})} \vvv{\bar{v}}_{i'j}  
\end{equation}
\begin{equation}\label{eq:3}
 \vvv{q}_{ij} = W^Q\bar{M}_{ij},~ \vvv{k}_{ij} = W^K\bar{M}_{ij},~  \vvv{v}_{ij} = W^V\bar{M}_{ij} \Rightarrow  F(M)_{ij}= \sum_{j'=1}^P \frac{\exp(\vvv{q}_{ij} \cdot \vvv{k}_{ij'})}{\sum_{l=1}^P \exp(\vvv{q}_{ij} \cdot \vvv{k}_{il})} \vvv{v}_{ij'}
\end{equation}

Here, $M_{i,j}\in \mathbb{R}^d$ are the features associated with the $j$-th point in the $i$-th frame. The attention mechanism defined in the first equation above \eqref{eq:1} is augmented with standard temporal positional encoding in the first layer and replaces the translation equivariant function $L_i$ applied to the columns of $M$ (Eq.\eqref{eq:DSS}). The attention in the second equation \eqref{eq:3} implements the set aggregation (summation) (also in Eq.\eqref{eq:DSS}) applied to the rows of $M$.   As commonly done, we use transformers with 16 attention heads  \cite{vaswani2017attention}.

\vspace{-1.5mm}
\subsection{Constraining 3D motion and camera poses via low-rank assumption}
\vspace{-1.5mm}
\label{section::output_parameterization}
Given our 2D tracks, we aim to characterize the motion of the points by decomposing them into the global camera motion and the 3D motion of objects in the scene. The 2D motion of static scene points provides useful constraints for estimating the camera motion. However, as previously mentioned, predicting camera and dynamic 3D motion solely from 2D motion is an ill-posed problem without additional constraints \cite{akhter2008nonrigid}. 
We tackle this challenge by adding two mechanisms to our architecture: (1) low-rank movement assumption; and (2) specific modeling of the static scene for camera estimation.

\paragraph{Low-rank movement assumption.} First, motivated by classical orthographic Non-Rigid Structure from Motion \cite{bregler2000recovering}, we constrain the output points to be formulated by a linear combination of input-specific basis elements. 
Specifically, given the input 2D point tracks,  $M\in \mathbb{R}^{N \times P \times 3}$, our network predicts $K$ point clouds: $B_1 (M),\dots, B_K(M)\in \mathbb{R}^{P \times 3}$ and $N(K-1)$  linear coefficients, $\{c_{1k}(M)\}_{k=2}^K,\dots\{c_{Nk(M)}\}_{k=2}^K$ such that
$X_i(M)=B_1(M)+\sum_{k=2}^K c_{ik}(M) B_k(M),$
where $X_i(M)\in \mathbb{R}^{P\times3}$ is the 3D point cloud at frame $i$. The point clouds and coefficients are computed by taking the output of the last equivariant layer as defined in the previous section and applying invariant aggregations on the respective dimension resulting in equivariant and invariant outputs. See more details in the appendix. We note that we deliberately chose the coefficient of $B_1(M)$ to be the constant $1$, the reason is explained in the next paragraph.

\paragraph{Specific modeling of the static scene for camera estimation.} Frequently, casual video data of dynamic scenes contains many static regions, which can be used to determine camera poses \cite{zhao2022particlesfm}. We leverage this observation by treating the first basis element $B_1(M) \in \mathbb{R}^{P \times 3}$ as a static approximation for all scene points and encourage $B_1(M)$ as well as the output camera poses to explain the 2D observations according to this approximation using a "static" reprojection loss ($\mathcal{L}_{\text{Static}}$, defined in the next section). 
We note, however, that a static point cloud is not likely to produce low reprojection errors for the non-static components, thus the reprojection error necessitates robustness to substantial errors from the non-static elements. To address this, our network predicts (equivariantly) $P$ motion level values $\gamma_1(M), \dots, \gamma_P(M)\in \mathbb{R}_+$, one for each point in our dynamic point cloud,  which we use to weight the reprojection errors from $B_1(M)$. The main idea is to give less weight to non-static points so that the static projection loss can disregard them.
Specifically, inspired by \cite{zhang2022structure}, each $\gamma_i(M)$ defines a Cauchy distribution that models the reprojection errors for its associated world point, such that a world point with higher $\gamma$ is expected to produce a wider error distribution. Empirically, as noted by \cite{zhang2022structure}, the Cauchy distribution tends to be more robust for modeling reprojection error uncertainties compared to Gaussian noise modeling \cite{kendall2017uncertainties}. Then, $\mathcal{L}_{\text{Static}}$, minimizes the negative log-likelihood under this assumption.    See details in Sec.~\ref{section::training}.

\vspace{-1.5mm}
\subsection{Training and losses}
\vspace{-1.5mm}
\label{section::training}
\textbf{Model outputs.} Given the input 2D point tracks $M\in \mathbb{R}^{N\times P\times 3}$, our network produces outputs as a function of $M$: linear bases and coefficients $B_1 (M),\dots, B_K(M)\in \mathbb{R}^{P \times 3},  \{c_{1k}(M)\}_{k=2}^K,\dots,\{c_{Nk}(M)\}_{k=2}^K\in \mathbb{R}$ which define a dynamic point cloud $X_1(M),\cdots,X_N(M) \in \mathbb{R}^{P \times 3}$,$  \gamma_1(M),\dots, \gamma_P(M)\in \mathbb{R}_+ $ movement level values,  and $ (R_1(M), \vvv{t}_1(M)), \dots ,(R_N(M), \vvv{t}_N(M)) \in SO(3)\times \mathbb{R}^3$ camera poses.

We use these network outputs to define an self-supervised loss function with respect to the current network weights and $M$ which is defined by:
\begin{equation}
    \mathcal{L}=\lambda_{\text{Reproject}} \mathcal{L}_{\text{Reproject}}+\lambda_{\text{Static}}\mathcal{L}_{\text{Static}}+\lambda_{\text{Negative}}\mathcal{L}_{\text{Negative}}+\lambda_{\text{Sparse}}\mathcal{L}_{\text{Sparse}}
\end{equation}

\paragraph{Reprojection loss.} The reprojection loss encourages the consistency between the output 3D point clouds and camera poses, to the input 2D observations:
\begin{equation}
       \mathcal{L}_{\text{Reproject}}=\frac{1}{\sum_{i=1}^N\sum_{j=1}^P M_{ij}^{o}} \sum_{i=1}^N\sum_{j=1}^P M_{ij}^{o} \mathcal{R}(X_{ij},R_i,\vvv{t}_i,M_{ij}^{xy})
\end{equation}
where $ \mathcal{R}(X_{ij},R_i,\vvv{t}_i,M_{ij}^{xy})$ is the reprojection error when projecting the point $X_{ij}$ with the camera pose $(R_i,\vvv{t}_i)$ with respect to the measured point $M_{ij}^{xy}$:
\begin{equation}
    \mathcal{R}(X_{ij},R_i,\vvv{t}_i,M_{ij}^{xy})=\norm{\frac{(R_i^T (\vvv{X}_{ij}-\vvv{t}_i))_{1,2}}{(R_i^T (\vvv{X}_{ij}-\vvv{t}_i))_{3}}-M_{ij}^{xy} }
\end{equation}

\paragraph{Static loss.} As discussed in Sec.~\ref{section::output_parameterization}, to better constrain the camera poses, the first predicted basis element $B_1(M)\in \mathbb{R}^{P \times 3}$ defines a static (fixed in time) point cloud approximation. Our network also predicts a movement coefficient $\gamma_j(M)$ for each world point that defines a zero-mean Cauchy distribution.  Given $\gamma_j$ and the reprojection error $r_{ij} =  \mathcal{R}(B_{1j},R_i,\vvv{t}_i,M_{ij}^{xy})$ of the $j^{th}$ point of $B_1$ that is projected by the $i^{th}$ camera, the negative log-likelihood of $r_{ij}$ distributed according to the $\gamma_j$-zero-mean Cauchy distribution is proportional to: 
\begin{equation}
    \mathcal{C}(r_{ij},\gamma_j) = \log\left(\gamma_j + \frac{r_{ij}^2}{\gamma_j}\right)
\end{equation}
Note, that this loss reduces the contribution of the reprojection errors for points with high $\gamma$, but also encourages $\gamma$ to be small, i.e. encouraging the static point cloud to represent the dynamic scene when possible.
Our static loss is the mean negative log-likelihood over all observed points in all frames:
\begin{equation}
           \mathcal{L}_{\text{Static}}=\frac{1}{\sum_{i=1}^N\sum_{j=1}^P M_{ij}^{o}} \sum_{i=1}^N\sum_{j=1}^P M_{ij}^{o}  \mathcal{C}(\mathcal{R}(B_{1j},R_i,\vvv{t}_i,M_{ij}^{xy}),\gamma_j) 
\end{equation}
\paragraph{Regularization losses.} As in \cite{moran2021deep} we regularize the observed points to be in front of the camera by:
\begin{equation}
    \mathcal{L}_{\text{Negative}}=-\sum_{i=1}^N\sum_{j=1}^P M_{ij}^{o} \text{ Min}( R_i^T (\vvv{X}_{ij}-\vvv{t}_i))_{3},0)
\end{equation}
We further find it beneficial to regularize the deviation from the static approximation $B_1$ to be sparse for static points, i.e. points with low $\gamma$ values:
\begin{equation}
    \mathcal{L}_{\text{Sparse}}=\frac{1}{P(K-1)}\sum_{k=2}^K\sum_{j=1}^P  \frac{1}{3\gamma_j} \left(| B_{kj1}|+|B_{kj2}|+|B_{kj3}|\right)
\end{equation}
where $\gamma$ is detached from the gradient calculation for this loss.

\begin{table}[t]\tiny
    \centering
    \caption{ \small \textbf{Pet evaluation}. \textbf{Top}: Baseline method results for structure or camera estimation (or both). \textbf{Bottom}: Our results with several configurations. (C),(D), or (CD) respectively indicate the object categories in the training set: cats, dogs, or both. BA and FT respectively indicate a post-processing of Bundle Adjustment or fine-tuning.   \label{tab::pets_test}  }
     \begin{tabular}{l |cc|cc|cc|cc|c|c|c|c|}
         \toprule
         \midrule
       & \multicolumn{2}{c}{Abs Rel $\downarrow$}  & \multicolumn{2}{|c|}{$\delta<1.25 \uparrow$}  & \multicolumn{2}{|c|}{$\delta<1.25^2 \uparrow$} & \multicolumn{2}{|c|}{$\delta<1.25^3 \uparrow$} & ATE $\downarrow$  & RPE Trans $\downarrow$  & RPE Rot $\downarrow$ & Time \\  
       &Dyn. & All&Dyn. & All&Dyn. & All&Dyn.  & All & (mm)& (mm)&(deg)& (min)\\
         \midrule
D-SLAM \cite{teed2021droid}&-&-&-&-&-&-&-&-&5.08&3.60&0.20&0.16 \\
ParticleSFM \cite{zhao2022particlesfm}&-&-&-&-&-&-&-&-&12.79&6.95&0.51&11.00 \\
RCVD \cite{kopf2021robust}&0.40&3.6E+07&0.43&0.72&0.75&0.90&0.92&0.96&43.95&25.77&2.31&20.00 \\
CasualSAM \cite{zhang2022structure}&\textbf{0.09}&\textbf{0.06}&\textbf{0.93}&\textbf{0.97}&0.99&\textbf{0.99}&\textbf{1.00}&\textbf{1.00}&6.90&3.95&0.22&1.3E+02 \\
MiDaS \cite{birkl2023midas}&0.16&6.2E+04&0.78&0.71&0.97&0.88&\textbf{1.00}&0.93&-&-&-&\textbf{0.15} \\
  \midrule 
 Ours (C)&0.11&0.08&0.88&0.92&0.99&0.98&\textbf{1.00}&\textbf{1.00}&8.96&3.79&0.23&\textbf{0.15} \\
Ours (C)+BA&0.11&0.08&0.88&0.92&0.99&0.98&\textbf{1.00}&\textbf{1.00}&4.22&2.86&0.17&\textbf{0.15} \\
Ours (C)+FT&\textbf{0.09}&\textbf{0.06}&0.90&0.96&\textbf{1.00}&\textbf{0.99}&\textbf{1.00}&\textbf{1.00}&4.00&\textbf{2.74}&\textbf{0.16}&4.86 \\
Ours (D)&0.12&0.08&0.85&0.91&0.99&\textbf{0.99}&\textbf{1.00}&\textbf{1.00}&8.03&3.54&0.23&\textbf{0.15} \\
Ours (D)+BA&0.12&0.08&0.85&0.91&0.99&\textbf{0.99}&\textbf{1.00}&\textbf{1.00}&4.19&2.83&0.17&\textbf{0.15} \\
Ours (D)+FT&\textbf{0.09}&\textbf{0.06}&0.88&0.96&\textbf{1.00}&\textbf{0.99}&\textbf{1.00}&\textbf{1.00}&\textbf{3.98}&\textbf{2.74}&\textbf{0.16}&4.86 \\
Ours (CD)&0.12&0.08&0.85&0.91&0.98&0.98&\textbf{1.00}&\textbf{1.00}&8.11&3.68&0.24&\textbf{0.15} \\
Ours (CD)+BA&0.12&0.08&0.85&0.91&0.98&0.98&\textbf{1.00}&\textbf{1.00}&4.21&2.86&0.17&\textbf{0.15} \\
Ours (CD)+FT&\textbf{0.09}&\textbf{0.06}&0.90&0.96&\textbf{1.00}&\textbf{0.99}&\textbf{1.00}&\textbf{1.00}&\textbf{3.98}&\textbf{2.74}&\textbf{0.16}&4.86 \\

    \bottomrule
    \end{tabular}
    
\end{table}

\begin{table}[t]\tiny
    \centering
     \caption{ \small \textbf{Out-of-training-domain evaluation} \label{tab::nvidia_tes}. Evaluation metrics on monocular videos from \cite{yoon2020novel}. The table has the same structure as Tab.~\ref{tab::pets_test}.   }
     \begin{tabular}{l |cc|cc|cc|cc|c|c|c|c|}
         \toprule
         \midrule
       & \multicolumn{2}{c}{Abs Rel $\downarrow$}  & \multicolumn{2}{|c|}{$\delta<1.25 \uparrow$}  & \multicolumn{2}{|c|}{$\delta<1.25^2 \uparrow$} & \multicolumn{2}{|c|}{$\delta<1.25^3 \uparrow$} & ATE $\downarrow$  & RPE Trans $\downarrow$  & RPE Rot $\downarrow$ & Time \\  
       &Dyn. & All&Dyn. & All&Dyn. & All&Dyn.  & All & (mm)& (mm)&(deg)& (min)\\
         \midrule
D-SLAM \cite{teed2021droid}&-&-&-&-&-&-&-&-&7.96&10.91&0.07&0.18 \\
ParticleSFM \cite{zhao2022particlesfm}&-&-&-&-&-&-&-&-&26.66&23.83&0.20&2.13 \\
RCVD \cite{kopf2021robust}&0.19&2.6E+05&0.69&0.75&0.95&0.95&0.96&0.98&1.6E+02&3.2E+02&3.43&7.00 \\
CasualSAM \cite{zhang2022structure}&\textbf{0.05}&\textbf{0.03}&0.95&\textbf{0.99}&\textbf{0.99}&\textbf{1.00}&\textbf{1.00}&\textbf{1.00}&\textbf{7.81}&\textbf{10.09}&\textbf{0.06}&22.00 \\
MiDaS \cite{birkl2023midas}&2.8E+04&2.7E+05&0.59&0.58&0.73&0.72&0.83&0.80&-&-&-&\textbf{0.02} \\
  \midrule 
 Ours (C)&0.08&0.06&0.89&0.95&\textbf{0.99}&0.99&0.99&\textbf{1.00}&32.06&47.99&0.45&0.04 \\
Ours (C)+BA&0.08&0.06&0.89&0.95&\textbf{0.99}&0.99&0.99&\textbf{1.00}&8.67&12.36&0.08&0.04 \\
Ours (C)+FT&0.07&\textbf{0.03}&0.94&0.98&\textbf{0.99}&\textbf{1.00}&\textbf{1.00}&\textbf{1.00}&7.98&11.64&0.08&0.59 \\
Ours (D)&0.08&0.07&0.92&0.93&\textbf{0.99}&0.98&0.99&\textbf{1.00}&33.77&51.64&0.61&0.04 \\
Ours (D)+BA&0.08&0.07&0.92&0.93&\textbf{0.99}&0.98&0.99&\textbf{1.00}&8.40&12.06&0.08&0.04 \\
Ours (D)+FT&\textbf{0.05}&\textbf{0.03}&\textbf{0.97}&\textbf{0.99}&\textbf{0.99}&\textbf{1.00}&0.99&\textbf{1.00}&8.15&11.88&0.09&0.59 \\
Ours (CD)&0.10&0.08&0.93&0.94&\textbf{0.99}&0.99&\textbf{1.00}&\textbf{1.00}&36.17&53.94&0.67&0.04 \\
Ours (CD)+BA&0.10&0.08&0.93&0.94&\textbf{0.99}&0.99&\textbf{1.00}&\textbf{1.00}&8.62&12.49&0.08&0.04 \\
Ours (CD)+FT&0.06&\textbf{0.03}&\textbf{0.97}&\textbf{0.99}&\textbf{0.99}&\textbf{1.00}&0.99&\textbf{1.00}&8.04&11.84&0.09&0.59 \\
    \bottomrule
    \end{tabular}
   \vspace{-1.5mm}
\end{table}

\begin{table}[h]\tiny
    \centering
        \caption{\textbf{Ablation study.} \small \label{tab::ablation} The contribution of different parts from our method. See details in the text. }
     \begin{tabular}{l |cc|cc|cc|cc|cc|c|c|c|}
         \toprule
         \midrule
       & \multicolumn{2}{c}{Abs Rel $\downarrow$}  & \multicolumn{2}{|c|}{$\delta<1.25 \uparrow$}  & \multicolumn{2}{|c|}{$\delta<1.25^2 \uparrow$} & \multicolumn{2}{|c|}{$\delta<1.25^3 \uparrow$} &\multicolumn{2}{|c|}{Rep.(pix.) $\downarrow$ }& ATE $\downarrow$  & RPE Trans $\downarrow$  & RPE Rot $\downarrow$ \\  
       &Dyn. & All&Dyn. & All&Dyn. & All&Dyn. & All &Dyn. & All & (mm)& (mm)&(deg)\\
         \midrule

     Set of Sets&0.27 &0.15 &0.60 &0.77 &0.87 &0.94 &0.97 &0.99 &9.86 &5.33 &16.87 &5.53 &0.39 
     \\No $ \mathcal{L}_{\text{Static}}$&0.77 &0.36 &0.25 &0.46 &0.48 &0.70 &0.68 &0.82 &\textbf{1.00} &\textbf{0.86} &96.20 &29.86 &0.99 
     \\No $\gamma$&0.22 &0.16 &0.66 &0.73 &0.93 &0.91 &0.99 &0.97 &4.54 &2.41 &13.91 &4.86 &0.29 \\
     K=30&0.14 &0.09 &0.81 &0.90 &0.97 &\textbf{0.98} &0.99 &0.99 &4.88 &2.78 &9.39 &\textbf{3.68} &\textbf{0.23} \\
     K=2&\textbf{0.11} &\textbf{0.08} &\textbf{0.88} &0.91 &0.98 &\textbf{0.98} &\textbf{1.00} &\textbf{1.00} &8.58 &3.56 &9.31 &3.86 &0.25 \\
     DSS &1.65 &0.58 &0.19 &0.35 &0.34 &0.60 &0.47 &0.74 &63.75 &70.60 &34.90 &22.63 &1.64 \\
     No $\mathcal{L}_{\text{Sparse}}$&0.17 &0.13 &0.79 &0.80 &0.95 &0.94 &\textbf{1.00} &0.99 &4.57 &2.73 &11.79 &7.99 &0.55 \\
      Full&\textbf{0.11} &\textbf{0.08} &\textbf{0.88} &\textbf{0.92} &\textbf{0.99} &\textbf{0.98} &\textbf{1.00 }&\textbf{1.00} &3.98 &1.97 &\textbf{8.96} &3.79 &\textbf{0.23} \\
    \bottomrule
    \end{tabular}
\vspace{-3mm}
\end{table}

\vspace{-2mm}
\section{Experiments}
\vspace{-2mm}
In this section, we conduct experiments to verify our proposed network's performance on real-world casual videos. We began by training the network on specific domains and then evaluated its accuracy and running time on unseen videos from both, training and unseen domains.

\textbf{Training procedure.}
We trained our network on the cat and dog partitions from the COP3D dataset \cite{sinha2023common}, which contains a diverse set of casual real-world videos of pets. Our networks were trained from scratch three times to test our generalization capability between semantic categories: once on the cat partition, once on the dog partition, and once on both partitions combined. Training technical details are provided in the appendix. We use $K=12$ bases in all our experiments (Sec.~\ref{section::output_parameterization}).

\textbf{Evaluation data.}  To assess our framework's performance on pet videos, we curated a new dataset\footnote{ While the COP3D dataset provides cameras that were extracted by COLMAP \cite{schonberger2016structure}, we note that this evaluation data is insufficient in our case. This is because the dynamic structure was captured as well in part of their reconstruction which indicates that its reconstruction might not be accurate. Furthermore, the coordinates system units of these reconstructions are unknown. Finally, this dataset does not have any depth map information for evaluating the dynamic structure.} consisting of 21 casual videos of dogs and cats, each video containing 50 frames. These videos were captured using an RGBD (RGB-Depth) sensor. The depth maps were used as ground truth for evaluating the reconstructed structure. We extracted the cameras by running COLMAP on the images while masking out the pet areas with dilatated masks provided by \cite{zou2024segment}. The cameras were scaled to millimeter units using the provided GT depth.  Note that our network did not see this test data during training and it was not used to tune our hyperparameters.

Additionally, to evaluate our method on out-of-domain evaluation data, we used the Nvidia Dynamic Scenes Dataset \cite{yoon2020novel}. Specifically,  while our network was trained on pet videos, this dataset contains other dynamic object types, e.g. human, balloon, truck, and umbrella, with a different camera motion type.  The dataset contains 8 dynamic scenes which are captured by 12 synchronized cameras, enabling accurate depth estimation which is treated as GT for evaluating monocular depth estimation. The ground truth camera poses were calculated by \cite{schonberger2016structure} with the synchronized multiview camera rig and the ground truth dynamic masks. Similarly to \cite{li2021neural} we simulated 8 monocular dynamic video sequences using the camera rig, each with 24 frames, and used them for evaluation.

\begin{figure*}[t!]
    \centering
  \includegraphics[width=0.90\textwidth]{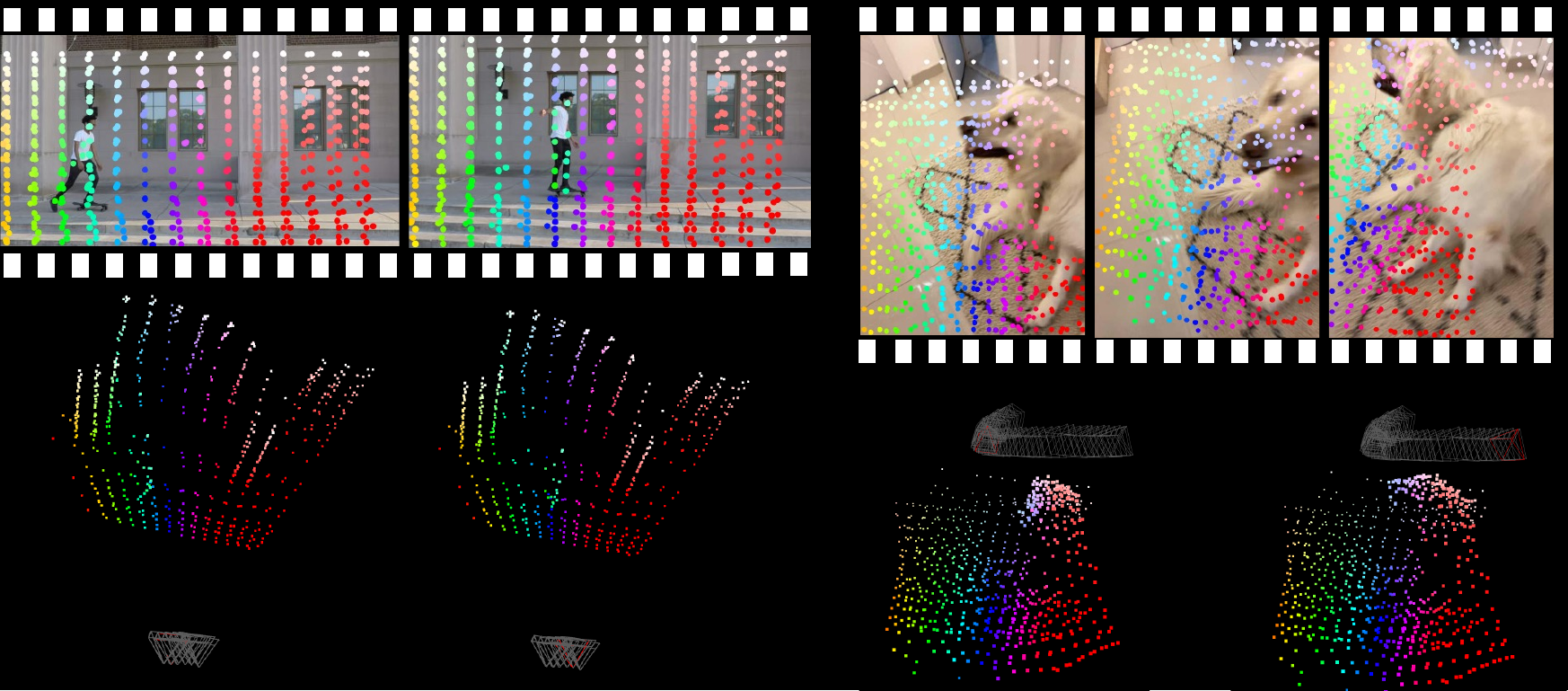}
  \caption{\small \textbf{Qualitative Results. }  \textbf{Top}. Frames from 2 different test video sequences with point tracks marked with corresponding colors. \textbf{Bottom}. A 3D visualization of our method's outputs, from two time stamps.  The camera trajectory is present as gray frustums, whereas the current camera is marked in red. The reconstructed 3D scene points are presented in corresponding colors to the input
tracks on the top.  The scene is observed from the same viewpoint, enabling the visualization of the dynamic reconstructed structure.  \vspace{-3mm} }
  \label{fig:qualitative}
\end{figure*}

\begin{figure*}[t!]
    \centering
  \includegraphics[width=0.99\textwidth]{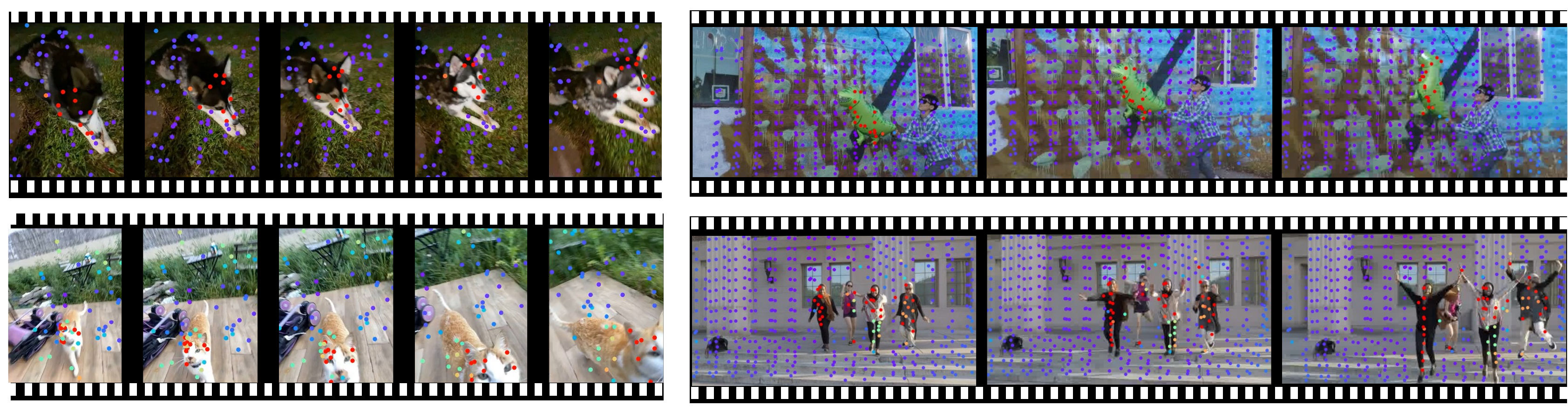}
  \caption{\small \label{fig:gamma}\textbf{$\gamma$ Visualization.} We show a visualization of the $\gamma$ outputs of our network that are described in Sec.~\ref{section::output_parameterization}. In each video sequence, we show the input tracks, where each color visualizes its movement level value, $\gamma$. Purple marks static points with low $\gamma$ whereas red marks dynamic points with high $\gamma$. Note, that our network did not get any direct supervision for these values, but only the raw point tracks predictions from \cite{karaev2023cotracker}. The $\gamma$ visualizations for cats were produced by the model that was only trained on dogs and vice versa. We note that our model generalizes well to out-of-domain (non-pet) cases as well. \vspace{-6mm}}
  
\end{figure*}

\vspace{-1.5mm}
\paragraph{Evaluation results.} 
Qualitative visualizations are presented in  Fig.~\ref{fig:qualitative}.\footnote{The reader is encouraged to watch the supplementary videos for better 4D perception. } We also show a visualization of the movement level values, $\gamma$ in Fig.~\ref{fig:gamma}. For comparisons, we chose state-of-the-art methods that as our method, can be applied to raw casual videos that were captured by standard pinhole camera models and do not need any static or dynamic segmentation. We evaluated both, the camera poses and the structure accuracies. Comparison results for the pet-test-set and out-of-domain dataset are presented in Tables \ref{tab::pets_test} and \ref{tab::nvidia_tes} respectively. The camera poses are evaluated compared to the GT, using the Absolute Translation Error(ATE), the Relative Translation Error(RTE), and the Relative Rotation Error(RRE) metrics after coordinates system alignment. We compare three training configurations of our method of training only on cats, only on dogs, and on both. As can be seen in the tables, the results are similar in all 3 cases.  Our output camera poses as inferred by the network are already accurate and outperform some of the prior methods. We further show the results of our method after a single and short round of Bundle Adjustment, which makes our method better than all baselines on the pet sequences, and comparable on the out-of-domain cases. 

Importantly, Tables \ref{tab::pets_test} and \ref{tab::nvidia_tes}  also compare the method's running times. It can be seen that our method, even with bundle adjustment, is the fastest camera prediction method. 
 Tables \ref{tab::pets_test} and \ref{tab::nvidia_tes}  also show structure evaluation with the depth evaluation metrics \cite{eigen2014depth} on the sampled point tracks. They demonstrate that our inferred structure is almost comparable to the state-of-the-art  \cite{zhang2022structure}  while taking significantly shorter running times (a few seconds for our method versus more than two hours for \cite{zhang2022structure} on pet videos). Short (0.6-5 minutes), per-sequence fine-tuning makes our method's accuracy comparable to \cite{zhang2022structure}, see appendix for details. In terms of running time, our method is a bit slower than MiDaS \cite{birkl2023midas}, which only provides depth maps without cameras, but achieves much better results.  We note that in contrast to the other methods that predict the dynamic depth, ours does not use any depth-from-single-image prior. 

\paragraph{Ablation study}
To evaluate the contribution of our different method parts we run an ablation study which is presented in Tab.~\ref{tab::ablation}. In this study, the training was always done on the cat partition from COP3D and evaluated on our test data which contains dogs and cats. First, we performed an ablation study on our transformer architecture by taking the architecture suggested by \cite{moran2021deep} ("Set of Sets") or the DSS architecture that uses only linear layers \cite{maron2020learning} ("DSS"). As the table shows our architecture ("Full") achieved significantly better results. 
To test the losses in our framework, we also evaluated the following: (1) ignoring the $\gamma$ outputs and using regular reprojection error on $B_1$ for all points ("No $\gamma$"); (2) removing our sparsity loss ("No $\mathcal{L}_{\text{Sparse}}$");   and (3) removing the static loss ("No $\mathcal{L}_{\text{Static}}$"). In all cases, the error increased whereas in the later one, the results became much worse.  
We further ablate the choice of $K=12$ as the number of linear bases, by trying 2 extreme numbers of $K=30,K=2$ (we saw no significant differences when we used nearby choices such as $K=11$). As can be seen in the table, when $K=30$ the output is not regularized enough and produces a higher depth error for the dynamic part. For $K=2$ the depth is regularized but the reprojection error  ("Rep. (pix.)") gets higher due to over-regularization. Overall, this study justifies our design choices ("Full").

\vspace{-2mm}
\section{Related Work}
\vspace{-2mm}

\paragraph{Simultaneous Localization and Mapping (SLAM) and Structure from Motion (SfM) } SfM pipelines seek to recover static 3D structure and camera poses from unordered images.\cite{tomasi1992shape,sturm1996factorization,schonberger2016structure,agarwal2011building,wu2013towards}. Learning-free pipelines \cite{schonberger2016structure} are effective but require repeated applications of  Bundle Adjustment (BA) \cite{triggs2000bundle}. \cite{moran2021deep,brynte2023learning} presented a method for learning prior from a dataset of multiview image sets, to accelerate SfM pipelines by using equivariant deep networks. Monocular Simultaneous Localization and Mapping (SLAM) methods \cite{mur2015orb,newcombe2011dtam,engel2014lsd,yang2020d3vo,bloesch2018codeslam,wang2021tartanvo,zhao2020towards,zhou2018deeptam,teed2018deepv2d} extract camera poses from video sequences while defining a scene map with keyframes. 
These methods assume static scenes, fail to produce the cameras in scenes with large portions of dynamic motion, and cannot reproduce dynamic parts of the scene. DROID-SLAM \cite{teed2021droid} used synthetic data with ground truth 3D supervision for learning to predict camera poses via deep-based BA on keyframes while excluding dynamic objects. ParticleSfM \cite{zhao2022particlesfm} 
filters out 2D dynamic content for reproducing the cameras in dynamic scenes, using its pre-trained motion prediction network. Both, \cite{teed2021droid,zhao2022particlesfm} do not infer the dynamic 3D structure. 
\vspace{-5mm}
\paragraph{Orthographic Non-Rigid SfM (NRSfM)} Bregler et al.  \cite{bregler2000recovering} introduced a factorization method for computing a non-rigid structure and rotation matrices from a point track matrix, by assuming a low dimensional basis model. While follow-up papers improved accuracies with different regularizations  \cite{kumar2022organic,dai2014simple,oskarsson2016trust,iglesias2020accurate} or neural representations \cite{novotny2019c3dpo,kong2019deep,sidhu2020neural}, the orthographic camera model assumption is in general not valid for casual videos. Furthermore, these methods often assume background subtraction as a preprocessing.  Even though a follow-up work proposed factorization solutions for pinhole cameras \cite{hartley2008perspective}, its sensitivity to noise \cite{jensen2021benchmark}, makes it impractical for casual videos. 

\paragraph{Test-time optimization for dynamic scenes } 
Recent methods \cite{luo2020consistent,kopf2021robust,zhang2022structure,zhang2021consistent} finetuned the monocular depth estimation from a pre-trained model \cite{Ranftl2022,Ranftl2021} using optical flow constraints  \cite{teed2020raft}, for obtaining consistent dense depth maps for a monocular video. \cite{zhang2022structure} further optimized motion maps for handling scenes with large dynamic motion.  \cite{xian2021space,gao2021dynamic} use depth from single image estimations, to improve novel view synthesis in dynamic scenes. \cite{liu2023robust} further optimizes for the unknown camera poses together with the dynamic radiance field optimization. \cite{parashar2017isometric,parashar2021robust} model a single deformable surface from a monocular video by applying isometric constraints.  LASR \cite{yang2021lasr}, ViSER \cite{yang2021viser} and BANMo \cite{yang2022banmo} optimize for a dynamic surface by assuming rigid bones and linear blend skinning weights.  However, all the above-mentioned methods require per-scene optimization, resulting in slow inference. Recently,  \cite{sinha2023common} presented the Common Pets in 3D (COP3D) dataset that contains casual, in-the-wild videos of pets, and used it to learn priors for novel view synthesis in dynamic scenes.

\paragraph{Point tracking } There has been a recent advance in 2D point tracking by learning  \cite{karaev2023cotracker,doersch2023tapir}, or optimization \cite{wang2023tracking} techniques. Concurrently, \cite{xiao2024spatialtracker} presented a method for jointly performing 2D tracking and 3D lifting, by learning to track with depth supervision while applying an as-rigid-as-possible loss. However, their method cannot predict camera poses or identify static parts directly.

\vspace{-2mm}
\section{Conclusions and limitations }
\vspace{-2mm}
We presented \ourmethod, a novel deep-learning framework that directly maps 2D motion tracks from casual videos into their corresponding dynamic structure and camera motion. Our approach features a deep learning architecture that considers the symmetries in the problem with designed intrinsic constraints to handle the ill-posed nature of this problem. Notably, our network was trained using only raw supervision of 2D point tracks extracted by an off-the-shelf method \cite{karaev2023cotracker} without any 3D supervision. Yet, it implicitly learned to predict camera poses and 3D structures while identifying the dynamic parts. During inference,  our method demonstrates significantly faster processing times compared to previous methods while achieving comparable accuracy. Furthermore, our network exhibits strong generalization capabilities, performing well even on semantic categories that were not present in the training data.

\paragraph{Limitations and future work.}
While our experiments demonstrated that our network is efficient, accurate, and capable of generalizing to unseen video categories, there are several limitations and future work directions that we would like to address. First, Our method cannot handle videos with too rapid motion, and in general, is limited by the accuracy of the tracking method \cite{karaev2023cotracker}. We note that any future improvements with point tracking in terms of accuracy and inference time will directly improve our method as well. Our method assumes enough motion parallax to constrain the depth values and fails to generate accurate camera poses without it. A future and interesting work would be to try combining depth-from-single-image prior as additional inputs to our network for handling cases with minimal motion parallax and improving accuracies.

\paragraph{Acknowledgments} HM is the Robert J. Shillman Fellow, and is supported by the Israel Science Foundation through a personal grant (ISF 264/23) and an equipment grant (ISF 532/23).

\bibliographystyle{splncs04}
\bibliography{main}

\begin{thebibliography}{10}
\providecommand{\url}[1]{\texttt{#1}}
\providecommand{\urlprefix}{URL }
\providecommand{\doi}[1]{https://doi.org/#1}

\bibitem{agarwal2011building}
Agarwal, S., Furukawa, Y., Snavely, N., Simon, I., Curless, B., Seitz, S.M., Szeliski, R.: Building rome in a day. Communications of the ACM  \textbf{54}(10),  105--112 (2011)

\bibitem{Agarwal_Ceres_Solver_2022}
Agarwal, S., Mierle, K., Team, T.C.S.: {Ceres Solver} (10 2023), \url{https://github.com/ceres-solver/ceres-solver}

\bibitem{akhter2008nonrigid}
Akhter, I., Sheikh, Y., Khan, S., Kanade, T.: Nonrigid structure from motion in trajectory space. Advances in neural information processing systems  \textbf{21} (2008)

\bibitem{birkl2023midas}
Birkl, R., Wofk, D., M{\"u}ller, M.: Midas v3.1 -- a model zoo for robust monocular relative depth estimation. arXiv preprint arXiv:2307.14460  (2023)

\bibitem{bloesch2018codeslam}
Bloesch, M., Czarnowski, J., Clark, R., Leutenegger, S., Davison, A.J.: Codeslam—learning a compact, optimisable representation for dense visual slam. In: Proceedings of the IEEE conference on computer vision and pattern recognition. pp. 2560--2568 (2018)

\bibitem{bregler2000recovering}
Bregler, C., Hertzmann, A., Biermann, H.: Recovering non-rigid 3d shape from image streams. In: Proceedings IEEE Conference on Computer Vision and Pattern Recognition. CVPR 2000 (Cat. No. PR00662). vol.~2, pp. 690--696. IEEE (2000)

\bibitem{bronstein2021geometric}
Bronstein, M.M., Bruna, J., Cohen, T., Veli{\v{c}}kovi{\'c}, P.: Geometric deep learning: Grids, groups, graphs, geodesics, and gauges. arXiv preprint arXiv:2104.13478  (2021)

\bibitem{brynte2023learning}
Brynte, L., Iglesias, J.P., Olsson, C., Kahl, F.: Learning structure-from-motion with graph attention networks. arXiv preprint arXiv:2308.15984  (2023)

\bibitem{dai2014simple}
Dai, Y., Li, H., He, M.: A simple prior-free method for non-rigid structure-from-motion factorization. International Journal of Computer Vision  \textbf{107},  101--122 (2014)

\bibitem{doersch2023tapir}
Doersch, C., Yang, Y., Vecerik, M., Gokay, D., Gupta, A., Aytar, Y., Carreira, J., Zisserman, A.: Tapir: Tracking any point with per-frame initialization and temporal refinement. arXiv preprint arXiv:2306.08637  (2023)

\bibitem{eigen2014depth}
Eigen, D., Puhrsch, C., Fergus, R.: Depth map prediction from a single image using a multi-scale deep network. Advances in neural information processing systems  \textbf{27} (2014)

\bibitem{engel2014lsd}
Engel, J., Sch{\"o}ps, T., Cremers, D.: Lsd-slam: Large-scale direct monocular slam. In: European conference on computer vision. pp. 834--849. Springer (2014)

\bibitem{gao2021dynamic}
Gao, C., Saraf, A., Kopf, J., Huang, J.B.: Dynamic view synthesis from dynamic monocular video. In: Proceedings of the IEEE/CVF International Conference on Computer Vision. pp. 5712--5721 (2021)

\bibitem{hartley2008perspective}
Hartley, R., Vidal, R.: Perspective nonrigid shape and motion recovery. In: Computer Vision--ECCV 2008: 10th European Conference on Computer Vision, Marseille, France, October 12-18, 2008, Proceedings, Part I 10. pp. 276--289. Springer (2008)

\bibitem{hartley2003multiple}
Hartley, R., Zisserman, A.: Multiple view geometry in computer vision. Cambridge university press (2003)

\bibitem{iglesias2020accurate}
Iglesias, J.P., Olsson, C., Valtonen~{\"O}rnhag, M.: Accurate optimization of weighted nuclear norm for non-rigid structure from motion. In: Computer Vision--ECCV 2020: 16th European Conference, Glasgow, UK, August 23--28, 2020, Proceedings, Part XXVII 16. pp. 21--37. Springer (2020)

\bibitem{jensen2021benchmark}
Jensen, S.H.N., Doest, M.E.B., Aan{\ae}s, H., Del~Bue, A.: A benchmark and evaluation of non-rigid structure from motion. International Journal of Computer Vision  \textbf{129}(4),  882--899 (2021)

\bibitem{karaev2023cotracker}
Karaev, N., Rocco, I., Graham, B., Neverova, N., Vedaldi, A., Rupprecht, C.: Cotracker: It is better to track together. arXiv preprint arXiv:2307.07635  (2023)

\bibitem{kendall2017uncertainties}
Kendall, A., Gal, Y.: What uncertainties do we need in bayesian deep learning for computer vision? Advances in neural information processing systems  \textbf{30} (2017)

\bibitem{kingma2014adam}
Kingma, D.P., Ba, J.: Adam: A method for stochastic optimization. arXiv preprint arXiv:1412.6980  (2014)

\bibitem{kong2019deep}
Kong, C., Lucey, S.: Deep non-rigid structure from motion. In: Proceedings of the IEEE/CVF International Conference on Computer Vision. pp. 1558--1567 (2019)

\bibitem{kopf2021robust}
Kopf, J., Rong, X., Huang, J.B.: Robust consistent video depth estimation. In: Proceedings of the IEEE/CVF Conference on Computer Vision and Pattern Recognition. pp. 1611--1621 (2021)

\bibitem{kumar2022organic}
Kumar, S., Van~Gool, L.: Organic priors in non-rigid structure from motion. In: European Conference on Computer Vision. pp. 71--88. Springer (2022)

\bibitem{li2021neural}
Li, Z., Niklaus, S., Snavely, N., Wang, O.: Neural scene flow fields for space-time view synthesis of dynamic scenes. In: Proceedings of the IEEE/CVF Conference on Computer Vision and Pattern Recognition. pp. 6498--6508 (2021)

\bibitem{liu2023robust}
Liu, Y.L., Gao, C., Meuleman, A., Tseng, H.Y., Saraf, A., Kim, C., Chuang, Y.Y., Kopf, J., Huang, J.B.: Robust dynamic radiance fields. In: Proceedings of the IEEE/CVF Conference on Computer Vision and Pattern Recognition. pp. 13--23 (2023)

\bibitem{luo2020consistent}
Luo, X., Huang, J.B., Szeliski, R., Matzen, K., Kopf, J.: Consistent video depth estimation. ACM Transactions on Graphics (ToG)  \textbf{39}(4),  71--1 (2020)

\bibitem{maron2020learning}
Maron, H., Litany, O., Chechik, G., Fetaya, E.: On learning sets of symmetric elements. In: International conference on machine learning. pp. 6734--6744. PMLR (2020)

\bibitem{mildenhall2021nerf}
Mildenhall, B., Srinivasan, P.P., Tancik, M., Barron, J.T., Ramamoorthi, R., Ng, R.: Nerf: Representing scenes as neural radiance fields for view synthesis. Communications of the ACM  \textbf{65}(1),  99--106 (2021)

\bibitem{moran2021deep}
Moran, D., Koslowsky, H., Kasten, Y., Maron, H., Galun, M., Basri, R.: Deep permutation equivariant structure from motion. In: Proceedings of the IEEE/CVF International Conference on Computer Vision. pp. 5976--5986 (2021)

\bibitem{mur2015orb}
Mur-Artal, R., Montiel, J.M.M., Tardos, J.D.: Orb-slam: a versatile and accurate monocular slam system. IEEE transactions on robotics  \textbf{31}(5),  1147--1163 (2015)

\bibitem{newcombe2011dtam}
Newcombe, R.A., Lovegrove, S.J., Davison, A.J.: Dtam: Dense tracking and mapping in real-time. In: 2011 international conference on computer vision. pp. 2320--2327. IEEE (2011)

\bibitem{novotny2019c3dpo}
Novotny, D., Ravi, N., Graham, B., Neverova, N., Vedaldi, A.: C3dpo: Canonical 3d pose networks for non-rigid structure from motion. In: Proceedings of the IEEE/CVF International Conference on Computer Vision. pp. 7688--7697 (2019)

\bibitem{oskarsson2016trust}
Oskarsson, M., Batstone, K., Astrom, K.: Trust no one: Low rank matrix factorization using hierarchical ransac. In: Proceedings of the IEEE Conference on Computer Vision and Pattern Recognition. pp. 5820--5829 (2016)

\bibitem{parashar2017isometric}
Parashar, S., Pizarro, D., Bartoli, A.: Isometric non-rigid shape-from-motion with riemannian geometry solved in linear time. IEEE transactions on pattern analysis and machine intelligence  \textbf{40}(10),  2442--2454 (2017)

\bibitem{parashar2021robust}
Parashar, S., Pizarro, D., Bartoli, A.: Robust isometric non-rigid structure-from-motion. IEEE Transactions on Pattern Analysis and Machine Intelligence  \textbf{44}(10),  6409--6423 (2021)

\bibitem{Ranftl2021}
Ranftl, R., Bochkovskiy, A., Koltun, V.: Vision transformers for dense prediction. ICCV  (2021)

\bibitem{Ranftl2022}
Ranftl, R., Lasinger, K., Hafner, D., Schindler, K., Koltun, V.: Towards robust monocular depth estimation: Mixing datasets for zero-shot cross-dataset transfer. IEEE Transactions on Pattern Analysis and Machine Intelligence  \textbf{44}(3) (2022)

\bibitem{schonberger2016structure}
Schonberger, J.L., Frahm, J.M.: Structure-from-motion revisited. In: Proceedings of the IEEE conference on computer vision and pattern recognition. pp. 4104--4113 (2016)

\bibitem{sidhu2020neural}
Sidhu, V., Tretschk, E., Golyanik, V., Agudo, A., Theobalt, C.: Neural dense non-rigid structure from motion with latent space constraints. In: Computer Vision--ECCV 2020: 16th European Conference, Glasgow, UK, August 23--28, 2020, Proceedings, Part XVI 16. pp. 204--222. Springer (2020)

\bibitem{sinha2023common}
Sinha, S., Shapovalov, R., Reizenstein, J., Rocco, I., Neverova, N., Vedaldi, A., Novotny, D.: Common pets in 3d: Dynamic new-view synthesis of real-life deformable categories. In: Proceedings of the IEEE/CVF Conference on Computer Vision and Pattern Recognition. pp. 4881--4891 (2023)

\bibitem{sturm1996factorization}
Sturm, P., Triggs, B.: A factorization based algorithm for multi-image projective structure and motion. In: Computer Vision—ECCV'96: 4th European Conference on Computer Vision Cambridge, UK, April 15--18, 1996 Proceedings Volume II 4. pp. 709--720. Springer (1996)

\bibitem{teed2018deepv2d}
Teed, Z., Deng, J.: Deepv2d: Video to depth with differentiable structure from motion. arXiv preprint arXiv:1812.04605  (2018)

\bibitem{teed2020raft}
Teed, Z., Deng, J.: Raft: Recurrent all-pairs field transforms for optical flow. In: Computer Vision--ECCV 2020: 16th European Conference, Glasgow, UK, August 23--28, 2020, Proceedings, Part II 16. pp. 402--419. Springer (2020)

\bibitem{teed2021droid}
Teed, Z., Deng, J.: Droid-slam: Deep visual slam for monocular, stereo, and rgb-d cameras. Advances in neural information processing systems  \textbf{34},  16558--16569 (2021)

\bibitem{tomasi1992shape}
Tomasi, C., Kanade, T.: Shape and motion from image streams under orthography: a factorization method. International journal of computer vision  \textbf{9},  137--154 (1992)

\bibitem{triggs2000bundle}
Triggs, B., McLauchlan, P.F., Hartley, R.I., Fitzgibbon, A.W.: Bundle adjustment—a modern synthesis. In: Vision Algorithms: Theory and Practice: International Workshop on Vision Algorithms Corfu, Greece, September 21--22, 1999 Proceedings. pp. 298--372. Springer (2000)

\bibitem{vaswani2017attention}
Vaswani, A., Shazeer, N., Parmar, N., Uszkoreit, J., Jones, L., Gomez, A.N., Kaiser, {\L}., Polosukhin, I.: Attention is all you need. Advances in neural information processing systems  \textbf{30} (2017)

\bibitem{wan2021encoder}
Wan, Z., Li, Z., Tian, M., Liu, J., Yi, S., Li, H.: Encoder-decoder with multi-level attention for 3d human shape and pose estimation. In: Proceedings of the IEEE/CVF International Conference on Computer Vision. pp. 13033--13042 (2021)

\bibitem{wang2023tracking}
Wang, Q., Chang, Y.Y., Cai, R., Li, Z., Hariharan, B., Holynski, A., Snavely, N.: Tracking everything everywhere all at once. In: Proceedings of the IEEE/CVF International Conference on Computer Vision. pp. 19795--19806 (2023)

\bibitem{wang2021tartanvo}
Wang, W., Hu, Y., Scherer, S.: Tartanvo: A generalizable learning-based vo. In: Conference on Robot Learning. pp. 1761--1772. PMLR (2021)

\bibitem{wu2013towards}
Wu, C.: Towards linear-time incremental structure from motion. In: 2013 International Conference on 3D Vision-3DV 2013. pp. 127--134. IEEE (2013)

\bibitem{xian2021space}
Xian, W., Huang, J.B., Kopf, J., Kim, C.: Space-time neural irradiance fields for free-viewpoint video. In: Proceedings of the IEEE/CVF Conference on Computer Vision and Pattern Recognition. pp. 9421--9431 (2021)

\bibitem{xiao2024spatialtracker}
Xiao, Y., Wang, Q., Zhang, S., Xue, N., Peng, S., Shen, Y., Zhou, X.: Spatialtracker: Tracking any 2d pixels in 3d space. arXiv preprint arXiv:2404.04319  (2024)

\bibitem{yang2021lasr}
Yang, G., Sun, D., Jampani, V., Vlasic, D., Cole, F., Chang, H., Ramanan, D., Freeman, W.T., Liu, C.: Lasr: Learning articulated shape reconstruction from a monocular video. In: Proceedings of the IEEE/CVF Conference on Computer Vision and Pattern Recognition. pp. 15980--15989 (2021)

\bibitem{yang2021viser}
Yang, G., Sun, D., Jampani, V., Vlasic, D., Cole, F., Liu, C., Ramanan, D.: Viser: Video-specific surface embeddings for articulated 3d shape reconstruction. Advances in Neural Information Processing Systems  \textbf{34},  19326--19338 (2021)

\bibitem{yang2022banmo}
Yang, G., Vo, M., Neverova, N., Ramanan, D., Vedaldi, A., Joo, H.: Banmo: Building animatable 3d neural models from many casual videos. In: Proceedings of the IEEE/CVF Conference on Computer Vision and Pattern Recognition. pp. 2863--2873 (2022)

\bibitem{yang2020d3vo}
Yang, N., Stumberg, L.v., Wang, R., Cremers, D.: D3vo: Deep depth, deep pose and deep uncertainty for monocular visual odometry. In: Proceedings of the IEEE/CVF conference on computer vision and pattern recognition. pp. 1281--1292 (2020)

\bibitem{yoon2020novel}
Yoon, J.S., Kim, K., Gallo, O., Park, H.S., Kautz, J.: Novel view synthesis of dynamic scenes with globally coherent depths from a monocular camera. In: Proceedings of the IEEE/CVF Conference on Computer Vision and Pattern Recognition. pp. 5336--5345 (2020)

\bibitem{zhang2022structure}
Zhang, Z., Cole, F., Li, Z., Rubinstein, M., Snavely, N., Freeman, W.T.: Structure and motion from casual videos. In: European Conference on Computer Vision. pp. 20--37. Springer (2022)

\bibitem{zhang2021consistent}
Zhang, Z., Cole, F., Tucker, R., Freeman, W.T., Dekel, T.: Consistent depth of moving objects in video. ACM Transactions on Graphics (TOG)  \textbf{40}(4),  1--12 (2021)

\bibitem{zhao2022particlesfm}
Zhao, W., Liu, S., Guo, H., Wang, W., Liu, Y.J.: Particlesfm: Exploiting dense point trajectories for localizing moving cameras in the wild. In: European Conference on Computer Vision. pp. 523--542. Springer (2022)

\bibitem{zhao2020towards}
Zhao, W., Liu, S., Shu, Y., Liu, Y.J.: Towards better generalization: Joint depth-pose learning without posenet. In: Proceedings of the IEEE/CVF Conference on Computer Vision and Pattern Recognition. pp. 9151--9161 (2020)

\bibitem{zhou2018deeptam}
Zhou, H., Ummenhofer, B., Brox, T.: Deeptam: Deep tracking and mapping. In: Proceedings of the European conference on computer vision (ECCV). pp. 822--838 (2018)

\bibitem{zhou2019continuity}
Zhou, Y., Barnes, C., Lu, J., Yang, J., Li, H.: On the continuity of rotation representations in neural networks. In: Proceedings of the IEEE/CVF conference on computer vision and pattern recognition. pp. 5745--5753 (2019)

\bibitem{zou2024segment}
Zou, X., Yang, J., Zhang, H., Li, F., Li, L., Wang, J., Wang, L., Gao, J., Lee, Y.J.: Segment everything everywhere all at once. Advances in Neural Information Processing Systems  \textbf{36} (2024)

\end{thebibliography}







\appendix

\section{Supplementary results}

\subsection{Video results}
We provide supplementary video outputs of several cases from our test set. Each video presents the input video frames with a set of pre-extracted point tracks that are used as input to our network and presented in corresponding colors (left side), and the output cameras and dynamic 3D structure (right side). The output camera
trajectory is presented as gray frustums, whereas the current camera is marked in red.
The reconstructed 3D scene points are presented in corresponding colors to the input
tracks. Note that the outputs presented in the videos were obtained at inference time, with a single feed-forward prediction, without any optimization or fine-tuning, on unseen test cases. 
\subsection{Extended quantitative evaluation }
Per-sequence and mean quantitative comparisons for our 21 pet test videos are presented in Tab.~\ref{tab::depth_per_seq}  and Tab.~\ref{tab::cam_per_seq}. Tables with similar structure for the out-of-domain dataset are presented in  Tab.~\ref{tab::depth_per_seq_nvidia}  and Tab.~\ref{tab::cam_per_seq_nvidia}.

  

%
%

\section{Implementation details}

\begin{table}[t]\small
    \centering
     \begin{tabular}{l|c |c|c|c|c}
         \toprule
         \midrule
        & Grid & ATE $\downarrow$  & RPE Trans $\downarrow$  & RPE Rot $\downarrow$ & Inference $\downarrow$  \\ 

       &size& (mm) &(mm)&(deg)&Time\\
         \midrule

        Ours (cats only) &15& 8.96&3.79&0.23& 0.16(+8.6) seconds  \\
        Ours (cats only)+BA &15& 4.22&2.86&0.17&0.40(+8.6) seconds\\

                Ours (cats only) &12& 9.18&3.81&0.24& 0.09(+7.8) seconds  \\
        Ours (cats only)+BA &12& 4.36&2.97&0.17&0.24(+7.8) seconds\\
        Ours (cats only) &10& 8.91&3.91&0.23& 0.05(+7.7) seconds  \\
        Ours (cats only)+BA &10& 4.44&3.01&0.18&0.16(+7.7) seconds\\

                 Ours (cats only) &7& 9.06&4.11&0.25& 0.02(+7.6) seconds  \\
        Ours (cats only)+BA &7& 4.93&3.52&0.20&0.08(+7.6) seconds\\

                 Ours (cats only) &5& 10.29&4.97&0.31& 0.01(+7.6) seconds  \\
        Ours (cats only)+BA &5& 8.08&6.33&0.38&0.05(+7.6) seconds\\

    \bottomrule
    \end{tabular}
    \caption{\label{tab::grid_size}\textbf{Tracking Grid Size Effect } Quantitative evaluation of the effect of reducing the number of sampled point tracks at inference time. We measure the camera pose accuracy and the running time. We also mention the point tracks extraction time in parenthesis (e.g. +8.6 seconds) which is performed by \cite{karaev2023cotracker} as a preprocess. As can be seen, our method can handle a smaller number of points but the accuracy slightly drops with fewer sampled points}
\end{table}

\subsection{Architecture technical details}
For learning high frequencies we map each input coordinate to sinusoidal functions as in \cite{mildenhall2021nerf} with $L=12$. We use 3 pairs of attention layers, each of frames attention followed by point attention.   Each point (after sinusoidal functions embedding) is mapped into $\mathbb{R}^{256}$ with a linear layer. Each attention layer is a function of the form 
$F:\mathbb{R}^{N\times P\times 256}\rightarrow  \mathbb{R}^{N\times P\times 256}$ (see details above).  Each attention uses 16 heads with $K,Q,V \in \mathbb{R}^{ (N \text{ or } P) \times 64}$ followed by a fully connected network with 1 hidden layer of $2048$ features. We then average over the rows to get per-point features $P_0 \in \mathbb{R}^{P\times 256}$ and over the columns to get per-frame features $F_0 \in \mathbb{R}^{N\times 256}$. Finally, we map $P_0$ to per-point outputs $P_1\in \mathbb{R}^{P\times (3K+1)}$ ( K basis points and $\gamma$)  with a linear layer, and $F_0$ into per-camera outputs   $F_1\in \mathbb{R}^{N\times (6+3+K-1)}$ ($6$ for the rotation parameters \cite{zhou2019continuity}, $3$ for 
 the camera center, and $K-1$ linear coefficients) using a convolutional layer with a kernel size of $31$.    
\subsection{Training details}

In total, we used $733$ cat videos and $753$ dog videos for training. We trained our networks for $7000$ and $3500$ epochs for the single-class and multi-class setups respectively. Training our method lasts about one week on a single Tesla V100 GPU with 32GB memory. We used the Adam optimizer \cite{kingma2014adam} with a learning rate of $10^{-4}$. Our method assumes known camera internal parameters which are provided by the dataset and used to normalize the point tracks as a preprocessing step.



\subsection{Other implementation details}
\paragraph{Point tracks sampling} For building $M\in \mathbb{R}^{N\times P \times 3}$ we use the implementation of \cite{karaev2023cotracker}. We sample a uniform grid of $15\times 15$ 2D points, starting from frame number $0,20,40,\dots$, and then track these points throughout the entire video (backward and forward). In Tab.~\ref{tab::grid_size} we show the effect of reducing the grid size at inference time, in terms of camera pose accuracy and running time. During training, at each iteration, we randomly sample 20-50 frames from the training videos and 100 point tracks, i.e. $20 \leq N\leq 50$ and $P=100$. When sampling cameras and point tracks of size $N\times P \times 3 $ from a larger tensor of size $N' \times P' \times 3$ we only take a point track if its starting tracking time is in the range $[t-\frac{N}{2}, t+\frac{3N}{2}]$, where t is the first sampled index.   At inference time we take all the available point tracks.  In both, training and inference time, we keep only point tracks that are observed in more than $10$ frames. 

\paragraph{Finetuning details} For our fine-tuning (FT) in the main paper, we applied per-sequence fine-tunning of 500,100 iterations starting from our final checkpoint, for pets,out-of-domain data respectively. The fine-tuning is done as a post-processing by minimizing the original loss function on the given test video. 
\paragraph{Test Set} We used the RGBD camera of the iPhone 11 to record our 21 test videos of dogs and cats. Each frame has a resolution of $640\times 480$ pixels. Note that the training set contained various types of resolutions. All pet owners who were photographed gave their permission for the animals to be photographed. For evaluation only, we define a point track as dynamic if its associated GT mask value is $1$ for at least $40$ frames. The GT masks are obtained by running \cite{zou2024segment} and searching for labels of dogs and cats. They were only used for evaluation and not used by our method at all. We verified that this data includes enough dynamic motion, by also including several videos that COLMAP failed to reconstruct without the masks. We further verified manually that the camera trajectories look reasonable.

The out-of-domain dataset contains video sequences with 24 frames, each of resolution of $546\times 288$. The GT dynamic masks are provided by the dataset. For this dataset, for evaluation only, we define a point track as dynamic if its associated GT mask value is $1$ for at least $15$ frames. 

\paragraph{Bundle Adjustment} After inference, as optional refinement,  we take the output static approximation $B_1\in \mathbb{R}^{P\times 3}$ and the output camera poses $\{R_1,\dots,R_N\}$, $\{\vvv{t}_1,\dots,\vvv{t}_N\}$ and apply Bundle Adjustment (BA). We use a 3D world point from $B_1$ if its associated $\gamma$ is below $0.008$ for the pets dataset and $0.005$ for out-of-domain dataset. We optimize reprojection errors of a given observation $M_{i,j}$,  only if it is observed, i.e. $M_{i,j}^{o}=1$ and if the initial reprojection error is below $10$ pixels. We use the BA implementation provided by \cite{moran2021deep}, which is based on the Ceres package \cite{Agarwal_Ceres_Solver_2022}.  
\paragraph{Running times} All inference running times were computed on a machine with NVIDIA RTX A6000 GPU and Intel(R) Core(TM) i7-9800X  3.80GHz CPU. Extracting point tracks with \cite{karaev2023cotracker} took 8.6 and 2.5 seconds for each video on the pet-videos and out-of-domain videos respectively and included in the running time tables as part of our method inference time.

\paragraph{Training technical details} In all training setups, we used: $\lambda_{\text{Reprojection}}=50.0$, $\lambda_{\text{Static}}=1.0$, $\lambda_{\text{Negative}}=1.0$, $\lambda_{\text{Sparse}}=0.001$. At the beginning of the training, we pre-train the camera poses to be located behind and facing the origin. This prevents cases in which the cameras are located in the middle of the initial point cloud s.t. many points have negative depths, which may result in bad convergence. More specifically, the pre-train loss is: $\mathcal{L}_{\text{Pretrain}} = \frac{1}{N}\sum_{i=1}^N \frac{1}{100}\norm{\vvv{t}_i-[0,0,-15]^T}^2+ \norm{R_i-I}_{F}^2 $.


The pretrain runs until convergence ($\mathcal{L}_{\text{Pretrain}} <10^{-4})$. During the main training, we detach gradients from $B_1$ and $(R_1,\vvv{t}_1)\dots,(R_N,\vvv{t}_N)$ for $\mathcal{L}_{\text{Reproject}}$ to stabilize the training. Until epoch $50$ we sample sequences of $N$ in the range $[20,22]$, and then we increase the range to $[20,50]$. 

\begin{tiny}

    \centering
     \begin{longtable}{l|l|l|c |c|c|c|c|c|c}
         \toprule
         \midrule
       & && RCVD \cite{kopf2021robust}& MiDaS\cite{birkl2023midas}  & CasualSAM\cite{zhang2022structure} &Ours (C\&D) &Ours (C\&D) FT& Our (C) &Our (C) FT \\ 

      \multirow{8}{*}{Seq0}  &  \multirow{2}{*}{Abs Rel$\downarrow$ }& Dyn &0.11& 0.12&\textbf{0.05}&\textbf{0.05}& 0.06&\textbf{0.05}&0.06  \\
  & & All&1.90E+08& 8.80E+05&0.11&\textbf{0.06}& \textbf{0.06}&\textbf{0.06}&\textbf{0.06}  \\
 \cline{2-10}  &  \multirow{2}{*}{$\delta<1.25\uparrow$}& Dyn &0.94& 0.87&\textbf{1.00}&0.99& 0.98&0.99&0.98  \\
 & & All&0.71& 0.74&0.96&\textbf{0.98}& 0.97&\textbf{0.98}&0.97 \\
 \cline{2-10} &  \multirow{2}{*}{$\delta<1.25^2\uparrow$}& Dyn&\textbf{1.00}& \textbf{1.00}&\textbf{1.00}&\textbf{1.00}& \textbf{1.00}&\textbf{1.00}&\textbf{1.00} \\
 & & All&0.85& 0.87&0.97&\textbf{1.00}& 0.99&\textbf{1.00}&0.99 \\
 \cline{2-10} &  \multirow{2}{*}{$\delta<1.25^3\uparrow$}& Dyn &\textbf{1.00}& \textbf{1.00}&\textbf{1.00}&\textbf{1.00}& \textbf{1.00}&\textbf{1.00}&\textbf{1.00}  \\
& & All&0.89& 0.91&0.97&\textbf{1.00}& \textbf{1.00}&\textbf{1.00}&\textbf{1.00}  \\
\cline{1-10} 
 \multirow{8}{*}{Seq1}  &  \multirow{2}{*}{Abs Rel$\downarrow$ }& Dyn &0.29& 0.16&0.16&0.14& 0.12&0.13&\textbf{0.11}  \\
  & & All&0.19& 0.18&0.09&0.09& \textbf{0.07}&0.09&\textbf{0.07}  \\
 \cline{2-10}  &  \multirow{2}{*}{$\delta<1.25\uparrow$}& Dyn &0.45& 0.75&0.81&0.81& 0.86&\textbf{0.90}&0.87  \\
 & & All&0.68& 0.76&0.92&0.90& 0.92&\textbf{0.94}&0.92 \\
 \cline{2-10} &  \multirow{2}{*}{$\delta<1.25^2\uparrow$}& Dyn&0.85& 0.99&0.89&\textbf{1.00}& \textbf{1.00}&0.99&\textbf{1.00} \\
 & & All&0.93& 0.95&0.96&\textbf{1.00}& \textbf{1.00}&\textbf{1.00}&\textbf{1.00} \\
 \cline{2-10} &  \multirow{2}{*}{$\delta<1.25^3\uparrow$}& Dyn &0.99& \textbf{1.00}&\textbf{1.00}&\textbf{1.00}& \textbf{1.00}&\textbf{1.00}&\textbf{1.00}  \\
& & All&\textbf{1.00}& 0.99&\textbf{1.00}&\textbf{1.00}& \textbf{1.00}&\textbf{1.00}&\textbf{1.00}  \\
\cline{1-10} 
 \multirow{8}{*}{Seq2}  &  \multirow{2}{*}{Abs Rel$\downarrow$ }& Dyn &0.54& 0.10&0.06&0.06& \textbf{0.03}&\textbf{0.03}&0.04  \\
  & & All&0.20& 0.55&\textbf{0.06}&0.07& \textbf{0.06}&\textbf{0.06}&\textbf{0.06}  \\
 \cline{2-10}  &  \multirow{2}{*}{$\delta<1.25\uparrow$}& Dyn &0.20& 0.94&\textbf{0.99}&\textbf{0.99}& \textbf{0.99}&\textbf{0.99}&\textbf{0.99}  \\
 & & All&0.66& 0.66&\textbf{0.98}&0.96& \textbf{0.98}&0.97&\textbf{0.98} \\
 \cline{2-10} &  \multirow{2}{*}{$\delta<1.25^2\uparrow$}& Dyn&0.52& \textbf{1.00}&0.99&\textbf{1.00}& \textbf{1.00}&\textbf{1.00}&\textbf{1.00} \\
 & & All&0.89& 0.75&\textbf{0.99}&\textbf{0.99}& \textbf{0.99}&\textbf{0.99}&\textbf{0.99} \\
 \cline{2-10} &  \multirow{2}{*}{$\delta<1.25^3\uparrow$}& Dyn &0.91& \textbf{1.00}&\textbf{1.00}&\textbf{1.00}& \textbf{1.00}&\textbf{1.00}&\textbf{1.00}  \\
& & All&0.98& 0.80&\textbf{0.99}&\textbf{0.99}& \textbf{0.99}&\textbf{0.99}&\textbf{0.99}  \\
\cline{1-10} 
 \multirow{8}{*}{Seq3}  &  \multirow{2}{*}{Abs Rel$\downarrow$ }& Dyn &0.79& 0.25&0.07&0.15& \textbf{0.05}&0.07&\textbf{0.05}  \\
  & & All&0.22& 0.24&\textbf{0.06}&0.09& \textbf{0.06}&0.08&\textbf{0.06}  \\
 \cline{2-10}  &  \multirow{2}{*}{$\delta<1.25\uparrow$}& Dyn &0.10& 0.53&0.97&0.95& \textbf{0.99}&0.97&0.98  \\
 & & All&0.76& 0.74&\textbf{0.97}&0.91& \textbf{0.97}&0.92&\textbf{0.97} \\
 \cline{2-10} &  \multirow{2}{*}{$\delta<1.25^2\uparrow$}& Dyn&0.27& 0.99&\textbf{1.00}&\textbf{1.00}& \textbf{1.00}&\textbf{1.00}&\textbf{1.00} \\
 & & All&0.85& 0.92&\textbf{0.99}&0.98& \textbf{0.99}&0.98&\textbf{0.99} \\
 \cline{2-10} &  \multirow{2}{*}{$\delta<1.25^3\uparrow$}& Dyn &0.59& \textbf{1.00}&\textbf{1.00}&\textbf{1.00}& \textbf{1.00}&\textbf{1.00}&\textbf{1.00}  \\
& & All&0.92& 0.95&\textbf{0.99}&\textbf{0.99}& \textbf{0.99}&\textbf{0.99}&\textbf{0.99}  \\
\cline{1-10} 
 \multirow{8}{*}{Seq4}  &  \multirow{2}{*}{Abs Rel$\downarrow$ }& Dyn &0.31& 0.08&\textbf{0.06}&0.27& 0.15&0.19&0.15  \\
  & & All&0.17& 0.27&\textbf{0.09}&0.12& 0.11&0.10&0.11  \\
 \cline{2-10}  &  \multirow{2}{*}{$\delta<1.25\uparrow$}& Dyn &0.44& 0.98&\textbf{1.00}&0.54& 0.75&0.65&0.77  \\
 & & All&0.80& 0.65&\textbf{0.96}&0.84& 0.92&0.89&0.92 \\
 \cline{2-10} &  \multirow{2}{*}{$\delta<1.25^2\uparrow$}& Dyn&0.87& \textbf{1.00}&\textbf{1.00}&0.88& \textbf{1.00}&0.99&\textbf{1.00} \\
 & & All&0.97& 0.90&0.98&0.98& 0.99&\textbf{1.00}&0.99 \\
 \cline{2-10} &  \multirow{2}{*}{$\delta<1.25^3\uparrow$}& Dyn &\textbf{1.00}& \textbf{1.00}&\textbf{1.00}&\textbf{1.00}& \textbf{1.00}&\textbf{1.00}&\textbf{1.00}  \\
& & All&\textbf{1.00}& 0.94&0.99&\textbf{1.00}& \textbf{1.00}&\textbf{1.00}&\textbf{1.00}  \\
\cline{1-10} 
 \multirow{8}{*}{Seq5}  &  \multirow{2}{*}{Abs Rel$\downarrow$ }& Dyn &0.09& 0.08&\textbf{0.05}&0.07& 0.07&0.06&0.07  \\
  & & All&0.12& 3.75E+05&\textbf{0.03}&0.08& 0.05&0.06&0.04  \\
 \cline{2-10}  &  \multirow{2}{*}{$\delta<1.25\uparrow$}& Dyn &0.97& 0.98&\textbf{1.00}&0.99& 0.98&0.99&0.98  \\
 & & All&0.86& 0.86&\textbf{0.98}&0.91& 0.97&0.96&\textbf{0.98} \\
 \cline{2-10} &  \multirow{2}{*}{$\delta<1.25^2\uparrow$}& Dyn&\textbf{1.00}& \textbf{1.00}&\textbf{1.00}&\textbf{1.00}& \textbf{1.00}&\textbf{1.00}&\textbf{1.00} \\
 & & All&0.97& 0.95&\textbf{1.00}&0.97& \textbf{1.00}&0.98&\textbf{1.00} \\
 \cline{2-10} &  \multirow{2}{*}{$\delta<1.25^3\uparrow$}& Dyn &\textbf{1.00}& \textbf{1.00}&\textbf{1.00}&\textbf{1.00}& \textbf{1.00}&\textbf{1.00}&\textbf{1.00}  \\
& & All&0.99& 0.98&\textbf{1.00}&0.99& \textbf{1.00}&\textbf{1.00}&\textbf{1.00}  \\
\cline{1-10} 
 \multirow{8}{*}{Seq6}  &  \multirow{2}{*}{Abs Rel$\downarrow$ }& Dyn &0.35& 0.10&0.04&0.05& \textbf{0.03}&0.04&\textbf{0.03}  \\
  & & All&0.14& 0.15&0.05&\textbf{0.04}& 0.05&0.05&0.05  \\
 \cline{2-10}  &  \multirow{2}{*}{$\delta<1.25\uparrow$}& Dyn &0.47& 0.94&0.98&\textbf{0.99}& \textbf{0.99}&\textbf{0.99}&\textbf{0.99}  \\
 & & All&0.83& 0.87&0.96&\textbf{0.98}& 0.96&0.97&0.96 \\
 \cline{2-10} &  \multirow{2}{*}{$\delta<1.25^2\uparrow$}& Dyn&0.65& 0.99&0.99&\textbf{1.00}& \textbf{1.00}&\textbf{1.00}&\textbf{1.00} \\
 & & All&0.91& 0.95&\textbf{0.99}&\textbf{0.99}& 0.98&\textbf{0.99}&0.98 \\
 \cline{2-10} &  \multirow{2}{*}{$\delta<1.25^3\uparrow$}& Dyn &0.99& 0.99&\textbf{1.00}&\textbf{1.00}& \textbf{1.00}&\textbf{1.00}&\textbf{1.00}  \\
& & All&0.99& 0.98&\textbf{1.00}&\textbf{1.00}& 0.99&\textbf{1.00}&0.99  \\
\cline{1-10} 
 \multirow{8}{*}{Seq7}  &  \multirow{2}{*}{Abs Rel$\downarrow$ }& Dyn &0.39& \textbf{0.08}&0.09&0.15& 0.11&0.12&0.10  \\
  & & All&0.22& 0.17&\textbf{0.06}&0.10& \textbf{0.06}&0.09&\textbf{0.06}  \\
 \cline{2-10}  &  \multirow{2}{*}{$\delta<1.25\uparrow$}& Dyn &0.45& \textbf{0.95}&0.92&0.84& 0.91&0.90&0.91  \\
 & & All&0.65& 0.80&\textbf{0.97}&0.86& 0.96&0.91&\textbf{0.97} \\
 \cline{2-10} &  \multirow{2}{*}{$\delta<1.25^2\uparrow$}& Dyn&0.68& \textbf{1.00}&0.99&0.98& 0.98&0.98&0.98 \\
 & & All&0.86& 0.92&\textbf{0.99}&0.98& \textbf{0.99}&0.98&\textbf{0.99} \\
 \cline{2-10} &  \multirow{2}{*}{$\delta<1.25^3\uparrow$}& Dyn &0.93& \textbf{1.00}&\textbf{1.00}&\textbf{1.00}& 0.99&\textbf{1.00}&0.99  \\
& & All&0.97& 0.98&\textbf{1.00}&0.99& \textbf{1.00}&0.99&\textbf{1.00}  \\
\cline{1-10} 
 \multirow{8}{*}{Seq8}  &  \multirow{2}{*}{Abs Rel$\downarrow$ }& Dyn &0.47& 0.23&\textbf{0.05}&0.09& \textbf{0.05}&0.09&\textbf{0.05}  \\
  & & All&0.20& 3.80E+04&\textbf{0.03}&0.08& 0.04&0.07&0.04  \\
 \cline{2-10}  &  \multirow{2}{*}{$\delta<1.25\uparrow$}& Dyn &0.18& 0.68&\textbf{0.99}&0.97& \textbf{0.99}&0.98&\textbf{0.99}  \\
 & & All&0.72& 0.64&\textbf{0.99}&0.89& \textbf{0.99}&0.97&\textbf{0.99} \\
 \cline{2-10} &  \multirow{2}{*}{$\delta<1.25^2\uparrow$}& Dyn&0.69& 0.97&\textbf{1.00}&\textbf{1.00}& \textbf{1.00}&\textbf{1.00}&\textbf{1.00} \\
 & & All&0.91& 0.84&\textbf{1.00}&0.99& \textbf{1.00}&\textbf{1.00}&\textbf{1.00} \\
 \cline{2-10} &  \multirow{2}{*}{$\delta<1.25^3\uparrow$}& Dyn &0.97& \textbf{1.00}&\textbf{1.00}&\textbf{1.00}& \textbf{1.00}&\textbf{1.00}&\textbf{1.00}  \\
& & All&0.99& 0.92&\textbf{1.00}&\textbf{1.00}& \textbf{1.00}&\textbf{1.00}&\textbf{1.00}  \\
\cline{1-10} 
 \multirow{8}{*}{Seq9}  &  \multirow{2}{*}{Abs Rel$\downarrow$ }& Dyn &0.86& 0.26&0.18&0.22& \textbf{0.17}&0.22&0.18  \\
  & & All&0.33& 0.20&\textbf{0.09}&0.21& \textbf{0.09}&0.22&\textbf{0.09}  \\
 \cline{2-10}  &  \multirow{2}{*}{$\delta<1.25\uparrow$}& Dyn &0.05& 0.59&\textbf{0.80}&0.46& 0.63&0.45&0.60  \\
 & & All&0.65& 0.74&\textbf{0.92}&0.55& 0.88&0.52&0.87 \\
 \cline{2-10} &  \multirow{2}{*}{$\delta<1.25^2\uparrow$}& Dyn&0.28& 0.85&0.95&0.91& \textbf{1.00}&0.85&\textbf{1.00} \\
 & & All&0.78& 0.91&0.98&0.89& \textbf{0.99}&0.86&0.98 \\
 \cline{2-10} &  \multirow{2}{*}{$\delta<1.25^3\uparrow$}& Dyn &0.68& \textbf{1.00}&\textbf{1.00}&\textbf{1.00}& \textbf{1.00}&\textbf{1.00}&\textbf{1.00}  \\
& & All&0.90& 0.98&\textbf{1.00}&0.97& \textbf{1.00}&0.97&0.99  \\
\cline{1-10} 
 \multirow{8}{*}{Seq10}  &  \multirow{2}{*}{Abs Rel$\downarrow$ }& Dyn &0.08& 0.06&\textbf{0.02}&\textbf{0.02}& \textbf{0.02}&\textbf{0.02}&\textbf{0.02}  \\
  & & All&0.11& 0.24&0.03&0.03& \textbf{0.02}&0.04&\textbf{0.02}  \\
 \cline{2-10}  &  \multirow{2}{*}{$\delta<1.25\uparrow$}& Dyn &0.99& \textbf{1.00}&\textbf{1.00}&\textbf{1.00}& \textbf{1.00}&\textbf{1.00}&\textbf{1.00}  \\
 & & All&0.92& 0.73&\textbf{1.00}&\textbf{1.00}& \textbf{1.00}&0.99&\textbf{1.00} \\
 \cline{2-10} &  \multirow{2}{*}{$\delta<1.25^2\uparrow$}& Dyn&\textbf{1.00}& \textbf{1.00}&\textbf{1.00}&\textbf{1.00}& \textbf{1.00}&\textbf{1.00}&\textbf{1.00} \\
 & & All&0.99& 0.87&\textbf{1.00}&\textbf{1.00}& \textbf{1.00}&\textbf{1.00}&\textbf{1.00} \\
 \cline{2-10} &  \multirow{2}{*}{$\delta<1.25^3\uparrow$}& Dyn &\textbf{1.00}& \textbf{1.00}&\textbf{1.00}&\textbf{1.00}& \textbf{1.00}&\textbf{1.00}&\textbf{1.00}  \\
& & All&\textbf{1.00}& 0.95&\textbf{1.00}&\textbf{1.00}& \textbf{1.00}&\textbf{1.00}&\textbf{1.00}  \\
\cline{1-10} 
 \multirow{8}{*}{Seq11}  &  \multirow{2}{*}{Abs Rel$\downarrow$ }& Dyn &0.34& 0.14&\textbf{0.06}&0.10& 0.07&0.10&0.07  \\
  & & All&0.17& 0.38&\textbf{0.05}&0.08& \textbf{0.05}&0.07&\textbf{0.05}  \\
 \cline{2-10}  &  \multirow{2}{*}{$\delta<1.25\uparrow$}& Dyn &0.47& 0.79&\textbf{0.92}&0.90& \textbf{0.92}&0.90&\textbf{0.92}  \\
 & & All&0.74& 0.67&\textbf{0.97}&0.90& 0.96&0.94&0.96 \\
 \cline{2-10} &  \multirow{2}{*}{$\delta<1.25^2\uparrow$}& Dyn&0.86& \textbf{1.00}&0.99&0.96& 0.98&0.97&0.98 \\
 & & All&0.94& 0.78&\textbf{0.99}&\textbf{0.99}& \textbf{0.99}&\textbf{0.99}&\textbf{0.99} \\
 \cline{2-10} &  \multirow{2}{*}{$\delta<1.25^3\uparrow$}& Dyn &0.94& \textbf{1.00}&\textbf{1.00}&0.99& \textbf{1.00}&0.99&\textbf{1.00}  \\
& & All&0.99& 0.86&\textbf{1.00}&\textbf{1.00}& \textbf{1.00}&\textbf{1.00}&\textbf{1.00}  \\
\cline{1-10} 
 \multirow{8}{*}{Seq12}  &  \multirow{2}{*}{Abs Rel$\downarrow$ }& Dyn &0.37& 0.16&0.05&0.10& 0.05&0.09&\textbf{0.04}  \\
  & & All&0.16& 0.13&\textbf{0.05}&0.08& 0.06&0.08&0.07  \\
 \cline{2-10}  &  \multirow{2}{*}{$\delta<1.25\uparrow$}& Dyn &0.19& 0.79&\textbf{0.98}&\textbf{0.98}& 0.97&\textbf{0.98}&0.97  \\
 & & All&0.73& 0.85&\textbf{0.98}&0.95& 0.94&0.95&0.94 \\
 \cline{2-10} &  \multirow{2}{*}{$\delta<1.25^2\uparrow$}& Dyn&0.97& \textbf{1.00}&\textbf{1.00}&\textbf{1.00}& \textbf{1.00}&\textbf{1.00}&\textbf{1.00} \\
 & & All&0.98& 0.98&\textbf{1.00}&0.99& 0.98&0.99&0.98 \\
 \cline{2-10} &  \multirow{2}{*}{$\delta<1.25^3\uparrow$}& Dyn &\textbf{1.00}& \textbf{1.00}&\textbf{1.00}&\textbf{1.00}& \textbf{1.00}&\textbf{1.00}&\textbf{1.00}  \\
& & All&\textbf{1.00}& \textbf{1.00}&\textbf{1.00}&\textbf{1.00}& \textbf{1.00}&\textbf{1.00}&\textbf{1.00}  \\
\cline{1-10} 
 \multirow{8}{*}{Seq13}  &  \multirow{2}{*}{Abs Rel$\downarrow$ }& Dyn &0.31& 0.29&\textbf{0.09}&0.15& 0.11&0.14&0.11  \\
  & & All&3.31E+08& 0.35&\textbf{0.06}&0.10& 0.07&0.09&0.07  \\
 \cline{2-10}  &  \multirow{2}{*}{$\delta<1.25\uparrow$}& Dyn &0.50& 0.50&0.93&0.82& 0.92&0.85&\textbf{0.95}  \\
 & & All&0.67& 0.69&\textbf{0.97}&0.91& 0.96&0.93&\textbf{0.97} \\
 \cline{2-10} &  \multirow{2}{*}{$\delta<1.25^2\uparrow$}& Dyn&0.81& 0.84&\textbf{1.00}&\textbf{1.00}& \textbf{1.00}&\textbf{1.00}&\textbf{1.00} \\
 & & All&0.81& 0.87&\textbf{0.99}&0.98& \textbf{0.99}&\textbf{0.99}&\textbf{0.99} \\
 \cline{2-10} &  \multirow{2}{*}{$\delta<1.25^3\uparrow$}& Dyn &\textbf{1.00}& \textbf{1.00}&\textbf{1.00}&\textbf{1.00}& \textbf{1.00}&\textbf{1.00}&\textbf{1.00}  \\
& & All&0.87& 0.93&\textbf{1.00}&0.99& 0.99&0.99&0.99  \\
\cline{1-10} 
 \multirow{8}{*}{Seq14}  &  \multirow{2}{*}{Abs Rel$\downarrow$ }& Dyn &0.18& 0.17&\textbf{0.03}&0.05& 0.04&0.04&0.04  \\
  & & All&0.18& 0.34&\textbf{0.03}&0.04& \textbf{0.03}&\textbf{0.03}&\textbf{0.03}  \\
 \cline{2-10}  &  \multirow{2}{*}{$\delta<1.25\uparrow$}& Dyn &0.79& 0.72&\textbf{0.99}&\textbf{0.99}& \textbf{0.99}&\textbf{0.99}&\textbf{0.99}  \\
 & & All&0.76& 0.55&\textbf{1.00}&0.99& 0.99&0.99&0.99 \\
 \cline{2-10} &  \multirow{2}{*}{$\delta<1.25^2\uparrow$}& Dyn&0.93& 0.97&\textbf{1.00}&\textbf{1.00}& \textbf{1.00}&\textbf{1.00}&\textbf{1.00} \\
 & & All&0.94& 0.83&\textbf{1.00}&\textbf{1.00}& \textbf{1.00}&\textbf{1.00}&\textbf{1.00} \\
 \cline{2-10} &  \multirow{2}{*}{$\delta<1.25^3\uparrow$}& Dyn &0.99& \textbf{1.00}&\textbf{1.00}&\textbf{1.00}& \textbf{1.00}&\textbf{1.00}&\textbf{1.00}  \\
& & All&0.99& 0.92&\textbf{1.00}&\textbf{1.00}& \textbf{1.00}&\textbf{1.00}&\textbf{1.00}  \\
\cline{1-10} 
 \multirow{8}{*}{Seq15}  &  \multirow{2}{*}{Abs Rel$\downarrow$ }& Dyn &1.22& 0.12&0.15&0.27& \textbf{0.10}&0.18&\textbf{0.10}  \\
  & & All&0.33& 0.39&\textbf{0.09}&0.15& \textbf{0.09}&0.18&\textbf{0.09}  \\
 \cline{2-10}  &  \multirow{2}{*}{$\delta<1.25\uparrow$}& Dyn &0.01& 0.77&0.95&0.62& 0.96&0.79&\textbf{0.97}  \\
 & & All&0.65& 0.65&\textbf{0.97}&0.80& 0.94&0.70&0.95 \\
 \cline{2-10} &  \multirow{2}{*}{$\delta<1.25^2\uparrow$}& Dyn&0.12& \textbf{1.00}&0.99&0.93& 0.99&0.96&0.99 \\
 & & All&0.81& 0.83&\textbf{0.99}&0.97& \textbf{0.99}&0.92&\textbf{0.99} \\
 \cline{2-10} &  \multirow{2}{*}{$\delta<1.25^3\uparrow$}& Dyn &0.46& \textbf{1.00}&0.99&\textbf{1.00}& \textbf{1.00}&\textbf{1.00}&\textbf{1.00}  \\
& & All&0.90& 0.89&\textbf{0.99}&\textbf{0.99}& \textbf{0.99}&\textbf{0.99}&\textbf{0.99}  \\
\cline{1-10} 
 \multirow{8}{*}{Seq16}  &  \multirow{2}{*}{Abs Rel$\downarrow$ }& Dyn &0.28& 0.12&\textbf{0.07}&0.14& 0.11&0.14&0.11  \\
  & & All&0.21& 0.21&\textbf{0.05}&0.06& 0.06&0.06&0.06  \\
 \cline{2-10}  &  \multirow{2}{*}{$\delta<1.25\uparrow$}& Dyn &0.42& 0.92&\textbf{0.98}&0.81& 0.88&0.80&0.90  \\
 & & All&0.66& 0.79&\textbf{0.99}&0.96& 0.97&0.96&0.97 \\
 \cline{2-10} &  \multirow{2}{*}{$\delta<1.25^2\uparrow$}& Dyn&0.98& \textbf{1.00}&\textbf{1.00}&\textbf{1.00}& \textbf{1.00}&\textbf{1.00}&\textbf{1.00} \\
 & & All&0.95& 0.93&\textbf{1.00}&\textbf{1.00}& \textbf{1.00}&\textbf{1.00}&\textbf{1.00} \\
 \cline{2-10} &  \multirow{2}{*}{$\delta<1.25^3\uparrow$}& Dyn &\textbf{1.00}& \textbf{1.00}&\textbf{1.00}&\textbf{1.00}& \textbf{1.00}&\textbf{1.00}&\textbf{1.00}  \\
& & All&\textbf{1.00}& 0.96&\textbf{1.00}&\textbf{1.00}& \textbf{1.00}&\textbf{1.00}&\textbf{1.00}  \\
\cline{1-10} 
 \multirow{8}{*}{Seq17}  &  \multirow{2}{*}{Abs Rel$\downarrow$ }& Dyn &0.35& 0.12&0.14&\textbf{0.11}& 0.12&\textbf{0.11}&0.12  \\
  & & All&0.23& 0.15&0.08&0.07& \textbf{0.06}&0.07&0.07  \\
 \cline{2-10}  &  \multirow{2}{*}{$\delta<1.25\uparrow$}& Dyn &0.40& 0.88&0.76&\textbf{0.93}& 0.89&0.91&0.88  \\
 & & All&0.60& 0.81&0.89&\textbf{0.94}& \textbf{0.94}&0.93&\textbf{0.94} \\
 \cline{2-10} &  \multirow{2}{*}{$\delta<1.25^2\uparrow$}& Dyn&0.82& \textbf{0.99}&0.98&\textbf{0.99}& 0.98&0.98&0.98 \\
 & & All&0.87& 0.94&0.98&0.98& \textbf{0.99}&\textbf{0.99}&\textbf{0.99} \\
 \cline{2-10} &  \multirow{2}{*}{$\delta<1.25^3\uparrow$}& Dyn &0.95& \textbf{1.00}&\textbf{1.00}&0.99& 0.99&0.99&0.99  \\
& & All&0.95& 0.98&\textbf{1.00}&0.99& \textbf{1.00}&\textbf{1.00}&\textbf{1.00}  \\
\cline{1-10} 
 \multirow{8}{*}{Seq18}  &  \multirow{2}{*}{Abs Rel$\downarrow$ }& Dyn &0.48& 0.21&0.10&0.12& \textbf{0.07}&0.11&\textbf{0.07}  \\
  & & All&0.14& 0.32&\textbf{0.05}&0.10& 0.09&0.08&0.09  \\
 \cline{2-10}  &  \multirow{2}{*}{$\delta<1.25\uparrow$}& Dyn &0.46& 0.60&0.93&0.87& \textbf{0.96}&0.88&\textbf{0.96}  \\
 & & All&0.86& 0.65&\textbf{0.98}&0.91& 0.96&0.95&0.96 \\
 \cline{2-10} &  \multirow{2}{*}{$\delta<1.25^2\uparrow$}& Dyn&0.55& 0.95&0.98&\textbf{1.00}& 0.99&0.99&0.99 \\
 & & All&0.91& 0.84&\textbf{0.99}&0.98& 0.98&0.98&0.98 \\
 \cline{2-10} &  \multirow{2}{*}{$\delta<1.25^3\uparrow$}& Dyn &0.84& \textbf{1.00}&0.99&\textbf{1.00}& \textbf{1.00}&\textbf{1.00}&\textbf{1.00}  \\
& & All&0.97& 0.92&\textbf{1.00}&0.99& 0.98&0.99&0.98  \\
\cline{1-10} 
 \multirow{8}{*}{Seq19}  &  \multirow{2}{*}{Abs Rel$\downarrow$ }& Dyn &0.33& 0.14&0.06&\textbf{0.04}& 0.05&0.05&0.05  \\
  & & All&0.15& 0.27&\textbf{0.03}&\textbf{0.03}& \textbf{0.03}&0.05&\textbf{0.03}  \\
 \cline{2-10}  &  \multirow{2}{*}{$\delta<1.25\uparrow$}& Dyn &0.28& 0.80&0.97&0.97& 0.97&\textbf{0.98}&0.97  \\
 & & All&0.81& 0.74&\textbf{0.99}&\textbf{0.99}& \textbf{0.99}&\textbf{0.99}&\textbf{0.99} \\
 \cline{2-10} &  \multirow{2}{*}{$\delta<1.25^2\uparrow$}& Dyn&0.91& 0.97&\textbf{1.00}&\textbf{1.00}& 0.99&0.99&0.99 \\
 & & All&0.98& 0.87&\textbf{1.00}&\textbf{1.00}& \textbf{1.00}&\textbf{1.00}&\textbf{1.00} \\
 \cline{2-10} &  \multirow{2}{*}{$\delta<1.25^3\uparrow$}& Dyn &\textbf{1.00}& 0.99&\textbf{1.00}&\textbf{1.00}& \textbf{1.00}&\textbf{1.00}&\textbf{1.00}  \\
& & All&\textbf{1.00}& 0.92&\textbf{1.00}&\textbf{1.00}& \textbf{1.00}&\textbf{1.00}&\textbf{1.00}  \\
\cline{1-10} 
 \multirow{8}{*}{Seq20}  &  \multirow{2}{*}{Abs Rel$\downarrow$ }& Dyn &0.33& 0.27&\textbf{0.21}&0.27& 0.29&0.26&0.30  \\
  & & All&2.42E+08& 0.45&\textbf{0.04}&0.05& 0.05&0.05&0.05  \\
 \cline{2-10}  &  \multirow{2}{*}{$\delta<1.25\uparrow$}& Dyn &0.34& 0.49&\textbf{0.61}&0.44& 0.37&0.49&0.34  \\
 & & All&0.50& 0.42&\textbf{0.97}&0.95& 0.95&0.95&0.95 \\
 \cline{2-10} &  \multirow{2}{*}{$\delta<1.25^2\uparrow$}& Dyn&0.93& 0.90&\textbf{1.00}&\textbf{1.00}& \textbf{1.00}&\textbf{1.00}&\textbf{1.00} \\
 & & All&0.77& 0.72&\textbf{1.00}&\textbf{1.00}& \textbf{1.00}&\textbf{1.00}&\textbf{1.00} \\
 \cline{2-10} &  \multirow{2}{*}{$\delta<1.25^3\uparrow$}& Dyn &0.99& 0.99&\textbf{1.00}&\textbf{1.00}& \textbf{1.00}&\textbf{1.00}&\textbf{1.00}  \\
& & All&0.83& 0.85&\textbf{1.00}&\textbf{1.00}& \textbf{1.00}&\textbf{1.00}&\textbf{1.00}  \\
\cline{1-10} 
 \multirow{8}{*}{Mean}  &  \multirow{2}{*}{Abs Rel$\downarrow$}& Dyn &0.40& 0.16&\textbf{0.09}&0.12& \textbf{0.09}&0.11&\textbf{0.09}  \\
  & & All&3.63E+07& 6.16E+04&\textbf{0.06}&0.08& \textbf{0.06}&0.08&\textbf{0.06}  \\
 \cline{2-10}  &  \multirow{2}{*}{$\delta<1.25\uparrow$}& Dyn &0.43& 0.78&\textbf{0.93}&0.85& 0.90&0.88&0.90  \\
 & & All&0.72& 0.71&\textbf{0.97}&0.91& 0.96&0.92&0.96 \\
 \cline{2-10} &  \multirow{2}{*}{$\delta<1.25^2\uparrow$}& Dyn&0.75& 0.97&0.99&0.98& \textbf{1.00}&0.99&\textbf{1.00} \\
 & & All&0.90& 0.88&\textbf{0.99}&0.98& \textbf{0.99}&0.98&\textbf{0.99} \\
 \cline{2-10} &  \multirow{2}{*}{$\delta<1.25^3\uparrow$}& Dyn &0.92& \textbf{1.00}&\textbf{1.00}&\textbf{1.00}& \textbf{1.00}&\textbf{1.00}&\textbf{1.00}  \\
& & All&0.96& 0.93&\textbf{1.00}&\textbf{1.00}& \textbf{1.00}&\textbf{1.00}&\textbf{1.00}  \\
\cline{1-10} 
 \multirow{8}{*}{STD}  &  \multirow{2}{*}{Abs Rel$\downarrow$}& Dyn &0.27& 0.07&\textbf{0.05}&0.08& 0.06&0.07&0.06  \\
  & & All&9.39E+07& 2.04E+05&\textbf{0.02}&0.04& \textbf{0.02}&0.04&\textbf{0.02}  \\
 \cline{2-10}  &  \multirow{2}{*}{$\delta<1.25\uparrow$}& Dyn &0.29& 0.17&\textbf{0.10}&0.18& 0.15&0.16&0.16  \\
 & & All&0.10& 0.11&\textbf{0.03}&0.10& \textbf{0.03}&0.11&\textbf{0.03} \\
 \cline{2-10} &  \multirow{2}{*}{$\delta<1.25^2\uparrow$}& Dyn&0.26& 0.05&0.02&0.03& \textbf{0.01}&0.03&\textbf{0.01} \\
 & & All&0.07& 0.07&\textbf{0.01}&0.02& \textbf{0.01}&0.03&\textbf{0.01} \\
 \cline{2-10} &  \multirow{2}{*}{$\delta<1.25^3\uparrow$}& Dyn &0.15& \textbf{0.00}&\textbf{0.00}&\textbf{0.00}& \textbf{0.00}&\textbf{0.00}&\textbf{0.00}  \\
& & All&0.05& 0.05&0.01&0.01& \textbf{0.00}&0.01&\textbf{0.00}  \\
\cline{1-10} 

    \bottomrule

      \caption{\label{tab::depth_per_seq}\textbf{Depth accuracy, pet test set. } We show a comparison to previous methods on the
predicted depth for the point trajectories compared to their GT depths. We compare 4 ways of running our method. Ours (C\&D), Ours (C): Our inference time outputs for a model that was trained on cats and dogs or only on cats respectively.   Ours (C\&D) FT, Ours (C) FT: The outputs of our model trained on cats and dogs or cats respectively, after fine-tuning our losses for each specific video. As can be seen, fine-tuning improves our accuracy even more. }
    \end{longtable}
  
\end{tiny}

\pagebreak

\begin{tiny}

    \centering

     \begin{longtable}{l|l|c |c|c|c|c|c|c}
         \toprule
         \midrule
       & & DROID-SLAM\cite{teed2021droid} & ParticleSfM\cite{zhao2022particlesfm}  & RCVD\cite{kopf2021robust} &CasualSAM\cite{zhang2022structure} &Ours  & Ours &Ours \\
       & &  &   &  & & (C) & (C)+BA & (C)+FT \\

         \midrule
\multirow{3}{*}{Seq0}   & ATE(mm) &\textbf{3.71}& 6.10&64.67&5.36& 5.60&5.43&4.42  \\
 & RPE T.(mm)&3.05& 3.22&26.92&3.13& 3.63&3.04&\textbf{2.89}  \\
 & RPE R.(deg.)&\textbf{0.14}& 0.18&2.53&0.16& 0.22&0.15&\textbf{0.14} \\
\cline{1-9} 
\multirow{3}{*}{Seq1}   & ATE(mm) &1.91& 3.83&38.44&10.28& 13.21&4.32&\textbf{1.76}  \\
 & RPE T.(mm)&\textbf{1.32}& 1.49&25.23&3.00& 2.65&1.72&\textbf{1.32}  \\
 & RPE R.(deg.)&0.18& 0.23&2.43&0.67& 0.29&0.19&\textbf{0.16} \\
\cline{1-9} 
\multirow{3}{*}{Seq2}   & ATE(mm) &3.13& 4.68&39.46&\textbf{2.13}& 4.57&2.78&2.53  \\
 & RPE T.(mm)&3.62& 5.82&27.27&\textbf{1.97}& 3.24&2.62&2.49  \\
 & RPE R.(deg.)&0.08& 0.21&1.89&\textbf{0.06}& 0.12&0.09&0.09 \\
\cline{1-9} 
\multirow{3}{*}{Seq3}   & ATE(mm) &5.13& \textbf{2.26}&29.55&2.49& 7.83&2.66&2.92  \\
 & RPE T.(mm)&5.76& \textbf{2.00}&24.18&2.31& 3.13&2.27&2.30  \\
 & RPE R.(deg.)&0.17& \textbf{0.07}&2.38&0.08& 0.13&\textbf{0.07}&\textbf{0.07} \\
\cline{1-9} 
\multirow{3}{*}{Seq4}   & ATE(mm) &2.59& 4.38&56.16&2.65& 4.21&2.38&\textbf{2.28}  \\
 & RPE T.(mm)&2.05& 2.07&21.36&\textbf{1.74}& 2.40&2.05&2.04  \\
 & RPE R.(deg.)&0.11& 0.11&1.05&\textbf{0.09}& 0.15&0.11&0.11 \\
\cline{1-9} 
\multirow{3}{*}{Seq5}   & ATE(mm) &1.07& 0.83&14.22&\textbf{0.53}& 1.44&0.79&0.75  \\
 & RPE T.(mm)&1.02& 0.74&6.51&\textbf{0.52}& 0.88&0.68&0.68  \\
 & RPE R.(deg.)&0.09& 0.07&1.63&\textbf{0.05}& 0.12&0.07&0.06 \\
\cline{1-9} 
\multirow{3}{*}{Seq6}   & ATE(mm) &\textbf{26.08}& 31.07&48.31&26.12& 27.10&28.28&28.98  \\
 & RPE T.(mm)&10.57& 11.01&24.21&\textbf{10.31}& 10.44&10.92&10.82  \\
 & RPE R.(deg.)&0.66& 0.67&4.03&\textbf{0.62}& 0.69&0.69&0.68 \\
\cline{1-9} 
\multirow{3}{*}{Seq7}   & ATE(mm) &2.25& 28.48&47.25&\textbf{1.78}& 4.53&2.39&2.25  \\
 & RPE T.(mm)&2.34& 6.13&23.81&\textbf{1.72}& 2.32&1.98&1.99  \\
 & RPE R.(deg.)&0.15& 0.59&2.29&\textbf{0.11}& 0.20&0.16&0.16 \\
\cline{1-9} 
\multirow{3}{*}{Seq8}   & ATE(mm) &1.23& 38.79&44.06&2.00& 4.78&\textbf{0.84}&0.89  \\
 & RPE T.(mm)&1.07& 50.57&25.46&1.24& 1.54&0.71&\textbf{0.70}  \\
 & RPE R.(deg.)&0.07& 4.78&1.73&0.09& 0.13&0.06&\textbf{0.05} \\
\cline{1-9} 
\multirow{3}{*}{Seq9}   & ATE(mm) &21.74& 18.52&43.45&34.93& 24.15&3.92&\textbf{3.22}  \\
 & RPE T.(mm)&9.04& 7.96&21.35&21.06& 6.89&2.77&\textbf{2.08}  \\
 & RPE R.(deg.)&0.62& 0.47&3.47&0.71& 0.37&0.12&\textbf{0.10} \\
\cline{1-9} 
\multirow{3}{*}{Seq10}   & ATE(mm) &1.47& 1.75&22.49&\textbf{1.40}& 3.03&1.42&1.46  \\
 & RPE T.(mm)&1.84& 1.15&24.22&\textbf{1.11}& 1.87&1.25&1.22  \\
 & RPE R.(deg.)&0.22& 0.12&2.18&\textbf{0.10}& 0.21&0.12&0.12 \\
\cline{1-9} 
\multirow{3}{*}{Seq11}   & ATE(mm) &1.71& 3.60&19.10&\textbf{1.32}& 3.12&1.66&1.49  \\
 & RPE T.(mm)&1.65& 1.54&16.34&\textbf{1.21}& 3.12&1.75&1.55  \\
 & RPE R.(deg.)&0.08& 0.08&1.80&\textbf{0.06}& 0.12&0.08&0.07 \\
\cline{1-9} 
\multirow{3}{*}{Seq12}   & ATE(mm) &\textbf{1.70}& 3.64&20.82&2.51& 6.23&3.61&2.54  \\
 & RPE T.(mm)&\textbf{2.24}& 2.63&18.50&2.74& 4.04&3.02&2.84  \\
 & RPE R.(deg.)&\textbf{0.09}& 0.12&2.03&0.12& 0.23&0.15&0.14 \\
\cline{1-9} 
\multirow{3}{*}{Seq13}   & ATE(mm) &1.23& 2.38&33.49&1.49& 4.10&\textbf{1.17}&1.28  \\
 & RPE T.(mm)&1.19& 1.40&17.33&\textbf{1.00}& 1.72&1.12&1.09  \\
 & RPE R.(deg.)&0.13& 0.14&2.12&\textbf{0.12}& 0.19&\textbf{0.12}&\textbf{0.12} \\
\cline{1-9} 
\multirow{3}{*}{Seq14}   & ATE(mm) &5.42& 5.15&1.05E+02&24.95& 6.09&5.06&\textbf{4.93}  \\
 & RPE T.(mm)&3.40& \textbf{3.38}&69.36&9.45& 4.07&3.57&3.41  \\
 & RPE R.(deg.)&\textbf{0.18}& 0.19&3.81&0.65& 0.26&0.20&0.19 \\
\cline{1-9} 
\multirow{3}{*}{Seq15}   & ATE(mm) &7.69& 61.06&36.70&7.40& 36.51&\textbf{4.98}&5.41  \\
 & RPE T.(mm)&7.95& 17.57&28.18&6.93& 10.19&5.86&\textbf{5.72}  \\
 & RPE R.(deg.)&0.22& 0.58&2.85&0.19& 0.30&0.16&\textbf{0.15} \\
\cline{1-9} 
\multirow{3}{*}{Seq16}   & ATE(mm) &5.04& 5.06&36.42&\textbf{3.69}& 4.52&4.11&3.81  \\
 & RPE T.(mm)&4.53& 4.54&20.11&\textbf{4.02}& 4.47&4.25&4.15  \\
 & RPE R.(deg.)&0.28& 0.29&2.94&\textbf{0.25}& 0.31&0.28&0.27 \\
\cline{1-9} 
\multirow{3}{*}{Seq17}   & ATE(mm) &\textbf{1.12}& 34.07&77.91&2.08& 6.52&2.51&2.86  \\
 & RPE T.(mm)&\textbf{1.15}& 12.30&39.64&1.18& 2.14&1.50&1.55  \\
 & RPE R.(deg.)&0.10& 1.11&2.20&\textbf{0.08}& 0.17&0.13&0.13 \\
\cline{1-9} 
\multirow{3}{*}{Seq18}   & ATE(mm) &\textbf{2.98}& 6.25&36.54&4.91& 7.73&4.13&3.86  \\
 & RPE T.(mm)&4.05& 5.21&24.78&3.89& 4.34&\textbf{3.62}&3.65  \\
 & RPE R.(deg.)&0.23& 0.29&1.67&\textbf{0.21}& 0.24&\textbf{0.21}&\textbf{0.21} \\
\cline{1-9} 
\multirow{3}{*}{Seq19}   & ATE(mm) &\textbf{1.45}& 2.23&42.95&1.82& 8.51&2.79&2.36  \\
 & RPE T.(mm)&1.65& 1.72&29.12&\textbf{1.44}& 3.23&2.21&2.01  \\
 & RPE R.(deg.)&0.11& 0.11&2.09&\textbf{0.09}& 0.25&0.16&0.15 \\
\cline{1-9} 
\multirow{3}{*}{Seq20}   & ATE(mm) &8.13& 4.37&66.03&4.95& 4.30&\textbf{3.43}&4.06  \\
 & RPE T.(mm)&6.20& 3.57&27.34&\textbf{3.00}& 3.30&3.05&3.06  \\
 & RPE R.(deg.)&0.36& 0.20&1.35&\textbf{0.18}& 0.20&\textbf{0.18}&\textbf{0.18} \\
\cline{1-9} 
\multirow{3}{*}{Mean}   & ATE(mm) &5.08& 12.79&43.95&6.90& 8.96&4.22&\textbf{4.00}  \\
 & RPE T.(mm)&3.60& 6.95&25.77&3.95& 3.79&2.86&\textbf{2.74}  \\
 & RPE R.(deg.)&0.20& 0.51&2.31&0.22& 0.23&0.17&\textbf{0.16} \\
\cline{1-9} 
\multirow{3}{*}{STD}   & ATE(mm) &6.63& 16.32&21.22&9.55& 9.06&\textbf{5.68}&5.87  \\
 & RPE T.(mm)&2.80& 10.88&11.80&4.74& 2.52&\textbf{2.22}&\textbf{2.22}  \\
 & RPE R.(deg.)&0.16& 1.01&0.76&0.22& \textbf{0.13}&\textbf{0.13}&\textbf{0.13} \\
\cline{1-9}

    \bottomrule
      \caption{\label{tab::cam_per_seq}\textbf{Camera poses accuracy for pets. } We show a comparison to previous methods on the predicted camera poses. We compare 3 ways of running our method. Ours (C): Our inference time outputs (total inference time of 0.16 seconds) for a model that was trained only on cats. Ours (C)+BA: Our inference time outputs, followed by a short Bundle Adjustment (total inference time of 0.4 seconds) for a model that was trained only on cats.   Ours (C) FT: The outputs of the model that was trained only on cats after fine-tuning our losses for each specific video (total running time of about 5 minutes). As can be seen, after BA, our results are the most accurate compared to the other methods, and fine-tuning improves our accuracy even more. }
    \end{longtable}
  
\end{tiny}

\begin{tiny}

    \centering
     \begin{longtable}{l|l|l|c |c|c|c|c}
         \toprule
         \midrule
       & && RCVD \cite{kopf2021robust}& MiDaS\cite{birkl2023midas}  & CasualSAM\cite{zhang2022structure} &Ours (C\&D) &Ours (C\&D) FT  \\ 
  \multirow{8}{*}{Balloon1}  &  \multirow{2}{*}{Abs Rel$\downarrow$ }& Dyn &0.21& 0.12&0.04&0.09& \textbf{0.03}  \\
  & & All&0.14& 0.34&0.02&0.06& \textbf{0.01}  \\
 \cline{2-8}  &  \multirow{2}{*}{$\delta<1.25\uparrow$}& Dyn &0.42& 0.89&\textbf{0.98}&\textbf{0.98}& \textbf{0.98}  \\
 & & All&0.62& 0.72&\textbf{0.99}&\textbf{0.99}& \textbf{0.99} \\
 \cline{2-8} &  \multirow{2}{*}{$\delta<1.25^2\uparrow$}& Dyn&\textbf{1.00}& 0.98&0.99&0.99& 0.99 \\
 & & All&\textbf{1.00}& 0.81&\textbf{1.00}&\textbf{1.00}& \textbf{1.00} \\
 \cline{2-8} &  \multirow{2}{*}{$\delta<1.25^3\uparrow$}& Dyn &\textbf{1.00}& 0.99&\textbf{1.00}&\textbf{1.00}& \textbf{1.00}  \\
& & All&\textbf{1.00}& 0.88&\textbf{1.00}&\textbf{1.00}& \textbf{1.00}  \\
\cline{1-8} 
 \multirow{8}{*}{Balloon2}  &  \multirow{2}{*}{Abs Rel$\downarrow$ }& Dyn &0.10& 2.21E+05&0.04&0.04& \textbf{0.03}  \\
  & & All&0.14& 4.87E+05&\textbf{0.01}&0.06& \textbf{0.01}  \\
 \cline{2-8}  &  \multirow{2}{*}{$\delta<1.25\uparrow$}& Dyn &0.97& 0.95&\textbf{1.00}&0.99& \textbf{1.00}  \\
 & & All&0.83& 0.76&\textbf{1.00}&0.97& \textbf{1.00} \\
 \cline{2-8} &  \multirow{2}{*}{$\delta<1.25^2\uparrow$}& Dyn&\textbf{1.00}& 0.99&\textbf{1.00}&\textbf{1.00}& \textbf{1.00} \\
 & & All&\textbf{1.00}& 0.86&\textbf{1.00}&\textbf{1.00}& \textbf{1.00} \\
 \cline{2-8} &  \multirow{2}{*}{$\delta<1.25^3\uparrow$}& Dyn &\textbf{1.00}& 0.99&\textbf{1.00}&\textbf{1.00}& \textbf{1.00}  \\
& & All&\textbf{1.00}& 0.93&\textbf{1.00}&\textbf{1.00}& \textbf{1.00}  \\
\cline{1-8} 
 \multirow{8}{*}{DynamicFace}  &  \multirow{2}{*}{Abs Rel$\downarrow$ }& Dyn &0.14& 0.55&\textbf{0.01}&0.15& \textbf{0.01}  \\
  & & All&0.05& 4.75E+04&\textbf{0.01}&0.06& \textbf{0.01}  \\
 \cline{2-8}  &  \multirow{2}{*}{$\delta<1.25\uparrow$}& Dyn &0.94& 0.03&\textbf{0.99}&0.98& 0.98  \\
 & & All&0.98& 0.67&\textbf{1.00}&\textbf{1.00}& \textbf{1.00} \\
 \cline{2-8} &  \multirow{2}{*}{$\delta<1.25^2\uparrow$}& Dyn&\textbf{1.00}& 0.04&\textbf{1.00}&0.98& \textbf{1.00} \\
 & & All&\textbf{1.00}& 0.82&\textbf{1.00}&\textbf{1.00}& \textbf{1.00} \\
 \cline{2-8} &  \multirow{2}{*}{$\delta<1.25^3\uparrow$}& Dyn &\textbf{1.00}& 0.18&\textbf{1.00}&\textbf{1.00}& \textbf{1.00}  \\
& & All&\textbf{1.00}& 0.86&\textbf{1.00}&\textbf{1.00}& \textbf{1.00}  \\
\cline{1-8} 
 \multirow{8}{*}{Jumping}  &  \multirow{2}{*}{Abs Rel$\downarrow$ }& Dyn &0.17& 0.43&\textbf{0.05}&0.07& 0.07  \\
  & & All&0.12& 0.59&\textbf{0.02}&0.05& 0.04  \\
 \cline{2-8}  &  \multirow{2}{*}{$\delta<1.25\uparrow$}& Dyn &0.77& 0.07&\textbf{0.99}&0.96& 0.96  \\
 & & All&0.86& 0.22&\textbf{0.99}&0.97& 0.97 \\
 \cline{2-8} &  \multirow{2}{*}{$\delta<1.25^2\uparrow$}& Dyn&\textbf{1.00}& 0.30&\textbf{1.00}&\textbf{1.00}& 0.99 \\
 & & All&\textbf{1.00}& 0.38&\textbf{1.00}&\textbf{1.00}& \textbf{1.00} \\
 \cline{2-8} &  \multirow{2}{*}{$\delta<1.25^3\uparrow$}& Dyn &\textbf{1.00}& 0.75&\textbf{1.00}&\textbf{1.00}& \textbf{1.00}  \\
& & All&\textbf{1.00}& 0.64&\textbf{1.00}&\textbf{1.00}& \textbf{1.00}  \\
\cline{1-8} 
 \multirow{8}{*}{Playground}  &  \multirow{2}{*}{Abs Rel$\downarrow$ }& Dyn &0.35& 0.52&\textbf{0.08}&0.16& 0.16  \\
  & & All&0.30& 7.67E+03&\textbf{0.07}&0.15& 0.08  \\
 \cline{2-8}  &  \multirow{2}{*}{$\delta<1.25\uparrow$}& Dyn &0.36& 0.49&\textbf{0.93}&0.64& 0.89  \\
 & & All&0.44& 0.59&\textbf{0.96}&0.78& 0.94 \\
 \cline{2-8} &  \multirow{2}{*}{$\delta<1.25^2\uparrow$}& Dyn&0.61& 0.72&\textbf{0.99}&0.98& 0.94 \\
 & & All&0.71& 0.78&\textbf{0.98}&0.91& 0.97 \\
 \cline{2-8} &  \multirow{2}{*}{$\delta<1.25^3\uparrow$}& Dyn &0.67& 0.83&\textbf{1.00}&0.98& 0.94  \\
& & All&0.82& 0.87&\textbf{0.99}&0.98& 0.98  \\
\cline{1-8} 
 \multirow{8}{*}{Skating}  &  \multirow{2}{*}{Abs Rel$\downarrow$ }& Dyn &0.16& 0.24&0.15&0.12& \textbf{0.10}  \\
  & & All&0.10& 1.09&\textbf{0.02}&0.05& 0.04  \\
 \cline{2-8}  &  \multirow{2}{*}{$\delta<1.25\uparrow$}& Dyn &0.89& 0.59&0.76&\textbf{0.93}& \textbf{0.93}  \\
 & & All&0.92& 0.29&0.99&\textbf{1.00}& 0.99 \\
 \cline{2-8} &  \multirow{2}{*}{$\delta<1.25^2\uparrow$}& Dyn&\textbf{1.00}& 0.95&0.97&0.97& 0.97 \\
 & & All&\textbf{1.00}& 0.42&\textbf{1.00}&\textbf{1.00}& \textbf{1.00} \\
 \cline{2-8} &  \multirow{2}{*}{$\delta<1.25^3\uparrow$}& Dyn &\textbf{1.00}& 0.97&\textbf{1.00}&\textbf{1.00}& 0.97  \\
& & All&\textbf{1.00}& 0.53&\textbf{1.00}&\textbf{1.00}& \textbf{1.00}  \\
\cline{1-8} 
 \multirow{8}{*}{Truck}  &  \multirow{2}{*}{Abs Rel$\downarrow$ }& Dyn &0.32& 0.13&\textbf{0.03}&0.14& 0.08  \\
  & & All&2.06E+06& 0.22&\textbf{0.03}&0.14& 0.05  \\
 \cline{2-8}  &  \multirow{2}{*}{$\delta<1.25\uparrow$}& Dyn &0.26& 0.81&0.99&0.94& \textbf{1.00}  \\
 & & All&0.50& 0.71&\textbf{1.00}&0.79& \textbf{1.00} \\
 \cline{2-8} &  \multirow{2}{*}{$\delta<1.25^2\uparrow$}& Dyn&0.99& 0.99&\textbf{1.00}&\textbf{1.00}& \textbf{1.00} \\
 & & All&0.88& 0.94&\textbf{1.00}&\textbf{1.00}& \textbf{1.00} \\
 \cline{2-8} &  \multirow{2}{*}{$\delta<1.25^3\uparrow$}& Dyn &\textbf{1.00}& \textbf{1.00}&\textbf{1.00}&\textbf{1.00}& \textbf{1.00}  \\
& & All&\textbf{1.00}& 0.99&\textbf{1.00}&\textbf{1.00}& \textbf{1.00}  \\
\cline{1-8} 
 \multirow{8}{*}{Umbrella}  &  \multirow{2}{*}{Abs Rel$\downarrow$ }& Dyn &0.08& 0.51&\textbf{0.02}&0.03& \textbf{0.02}  \\
  & & All&0.10& 1.62E+06&\textbf{0.02}&0.05& \textbf{0.02}  \\
 \cline{2-8}  &  \multirow{2}{*}{$\delta<1.25\uparrow$}& Dyn &0.91& 0.87&\textbf{1.00}&\textbf{1.00}& \textbf{1.00}  \\
 & & All&0.84& 0.68&\textbf{1.00}&0.99& \textbf{1.00} \\
 \cline{2-8} &  \multirow{2}{*}{$\delta<1.25^2\uparrow$}& Dyn&\textbf{1.00}& 0.91&\textbf{1.00}&\textbf{1.00}& \textbf{1.00} \\
 & & All&\textbf{1.00}& 0.71&\textbf{1.00}&\textbf{1.00}& \textbf{1.00} \\
 \cline{2-8} &  \multirow{2}{*}{$\delta<1.25^3\uparrow$}& Dyn &\textbf{1.00}& 0.92&\textbf{1.00}&\textbf{1.00}& \textbf{1.00}  \\
& & All&\textbf{1.00}& 0.72&\textbf{1.00}&\textbf{1.00}& \textbf{1.00}  \\
\cline{1-8} 
 \multirow{8}{*}{Mean}  &  \multirow{2}{*}{Abs Rel$\downarrow$}& Dyn &0.19& 2.76E+04&\textbf{0.05}&0.10& 0.06  \\
  & & All&2.58E+05& 2.70E+05&\textbf{0.03}&0.08& \textbf{0.03}  \\
 \cline{2-8}  &  \multirow{2}{*}{$\delta<1.25\uparrow$}& Dyn &0.69& 0.59&0.95&0.93& \textbf{0.97}  \\
 & & All&0.75& 0.58&\textbf{0.99}&0.94& \textbf{0.99} \\
 \cline{2-8} &  \multirow{2}{*}{$\delta<1.25^2\uparrow$}& Dyn&0.95& 0.73&\textbf{0.99}&\textbf{0.99}& \textbf{0.99} \\
 & & All&0.95& 0.72&\textbf{1.00}&0.99& \textbf{1.00} \\
 \cline{2-8} &  \multirow{2}{*}{$\delta<1.25^3\uparrow$}& Dyn &0.96& 0.83&\textbf{1.00}&\textbf{1.00}& 0.99 \\
& & All&0.98& 0.80&\textbf{1.00}&\textbf{1.00}& \textbf{1.00}  \\
\cline{1-8} 
 \multirow{8}{*}{STD}  &  \multirow{2}{*}{Abs Rel$\downarrow$}& Dyn &0.10& 7.81E+04&\textbf{0.05}&\textbf{0.05}& \textbf{0.05} \\
  & & All&7.28E+05& 5.69E+05&\textbf{0.02}&0.04& \textbf{0.02} \\
 \cline{2-8}  &  \multirow{2}{*}{$\delta<1.25\uparrow$}& Dyn &0.29& 0.37&0.08&0.12& \textbf{0.04}  \\
 & & All&0.20& 0.21&\textbf{0.01}&0.10& 0.02 \\
 \cline{2-8} &  \multirow{2}{*}{$\delta<1.25^2\uparrow$}& Dyn&0.14& 0.37&\textbf{0.01}&\textbf{0.01}& 0.02 \\
 & & All&0.10& 0.21&\textbf{0.01}&0.03& \textbf{0.01} \\
 \cline{2-8} &  \multirow{2}{*}{$\delta<1.25^3\uparrow$}& Dyn &0.12& 0.28&\textbf{0.00}&0.01& 0.02  \\
& & All&0.06& 0.16&\textbf{0.00}&0.01& 0.01  \\
\cline{1-8} 

    \bottomrule

      \caption{\label{tab::depth_per_seq_nvidia}\textbf{Depth accuracy for out-of-domain data \cite{yoon2020novel}.   } We show a comparison to previous methods on the
predicted depth for the point trajectories compared to their GT depths. We compare 2 ways of running our method. Ours (C\&D): Our inference time outputs for a model that was trained on cats and dogs.   Ours (C\&D) FT: The outputs of our model trained on cats and dogs, after fine-tuning our losses for each specific video. As can be seen, fine-tuning improves our accuracy even more. }
    \end{longtable}
  
\end{tiny}

\begin{tiny}

    \centering

     \begin{longtable}{l|l|c |c|c|c|c|c|c}
         \toprule
         \midrule
       & & DROID-SLAM\cite{teed2021droid} & ParticleSfM\cite{zhao2022particlesfm}  & RCVD\cite{kopf2021robust} &CasualSAM\cite{zhang2022structure} &Ours  & Ours &Ours \\
       & &  &   &  & & (C) & (C)+BA & (C)+FT \\

   \multirow{3}{*}{Balloon1}   & ATE(mm) &\textbf{2.87}& 5.81&1.7E+02&5.57& 21.47&4.17&4.14  \\
 & RPE T.(mm)&\textbf{4.58}& 6.51&2.5E+02&4.96& 27.73&6.76&6.76  \\
 & RPE R.(deg.)&\textbf{0.05}& 0.08&3.35&\textbf{0.05}& 0.40&0.07&0.07 \\
\cline{1-9} 
\multirow{3}{*}{Balloon2}   & ATE(mm) &7.81& 13.51&3.5E+02&\textbf{7.74}& 41.25&10.22&9.92  \\
 & RPE T.(mm)&12.82& 14.16&3.9E+02&\textbf{11.47}& 77.12&16.80&16.78  \\
 & RPE R.(deg.)&0.13& 0.13&3.21&\textbf{0.11}& 0.84&0.17&0.17 \\
\cline{1-9} 
\multirow{3}{*}{DynamicFace}   & ATE(mm) &\textbf{2.80}& 9.71&1.1E+02&3.59& 32.79&4.11&3.73  \\
 & RPE T.(mm)&\textbf{1.71}& 7.93&2.4E+02&2.05& 48.39&3.20&3.16  \\
 & RPE R.(deg.)&\textbf{0.04}& 0.17&3.34&0.05& 1.04&0.07&0.06 \\
\cline{1-9} 
\multirow{3}{*}{Jumping}   & ATE(mm) &\textbf{7.65}& 13.31&2.8E+02&7.74& 24.24&8.38&8.61  \\
 & RPE T.(mm)&10.25& 11.27&2.8E+02&\textbf{8.69}& 36.34&11.35&12.13  \\
 & RPE R.(deg.)&\textbf{0.05}& 0.06&3.09&\textbf{0.05}& 0.21&0.07&0.08 \\
\cline{1-9} 
\multirow{3}{*}{Playground}   & ATE(mm) &7.62& 85.47&1.1E+02&5.45& 27.44&6.47&\textbf{5.06}  \\
 & RPE T.(mm)&9.51& 90.10&3.3E+02&7.68& 40.28&8.00&\textbf{6.42}  \\
 & RPE R.(deg.)&\textbf{0.10}& 0.75&4.69&0.11& 0.40&\textbf{0.10}&\textbf{0.10} \\
\cline{1-9} 
\multirow{3}{*}{Skating}   & ATE(mm) &\textbf{7.21}& 19.37&78.24&7.28& 27.57&9.21&8.88  \\
 & RPE T.(mm)&\textbf{8.64}& 24.76&3.2E+02&8.65& 45.02&11.19&11.44  \\
 & RPE R.(deg.)&\textbf{0.04}& 0.15&3.91&0.05& 0.24&0.07&0.07 \\
\cline{1-9} 
\multirow{3}{*}{Truck}   & ATE(mm) &22.55& -&1.1E+02&17.70& 42.47&19.49&\textbf{17.53}  \\
 & RPE T.(mm)&31.68& -&3.6E+02&30.24& 69.22&34.61&\textbf{28.37}  \\
 & RPE R.(deg.)&0.06& -&2.77&\textbf{0.05}& 0.26&\textbf{0.05}&\textbf{0.05} \\
\cline{1-9} 
\multirow{3}{*}{Umbrella}   & ATE(mm) &\textbf{5.20}& 39.45&66.01&7.38& 39.27&7.33&5.99  \\
 & RPE T.(mm)&8.11& 12.08&3.7E+02&7.01& 39.85&\textbf{6.98}&8.03  \\
 & RPE R.(deg.)&0.04& 0.05&3.05&\textbf{0.03}& 0.20&\textbf{0.03}&\textbf{0.03} \\
\cline{1-9} 
\multirow{3}{*}{Mean}   & ATE(mm) &7.96& 26.66&1.6E+02&\textbf{7.81}& 32.06&8.67&7.98  \\
 & RPE T.(mm)&10.91& 23.83&3.2E+02&\textbf{10.09}& 47.99&12.36&11.64  \\
 & RPE R.(deg.)&0.07& 0.20&3.43&\textbf{0.06}& 0.45&0.08&0.08 \\
\cline{1-9} 
\multirow{3}{*}{STD}   & ATE(mm) &6.25& 28.13&1.0E+02&\textbf{4.25}& 8.11&4.89&4.50  \\
 & RPE T.(mm)&9.06& 29.82&55.63&8.60& 16.82&9.86&\textbf{7.95}  \\
 & RPE R.(deg.)&\textbf{0.03}& 0.25&0.61&\textbf{0.03}& 0.32&0.04&0.04 \\
\cline{1-9}

    \bottomrule
      \caption{\label{tab::cam_per_seq_nvidia}\textbf{Camera poses accuracy for out-of-domain data \cite{yoon2020novel}.   } We show a comparison to previous methods on the predicted camera poses. We compare 3 ways of running our method. Ours (C): Our inference time outputs. Ours (C)+BA: Our inference time outputs, followed by a short Bundle Adjustment for a model that was trained only on cats.   Ours (C) FT: The outputs of the model that was trained only on cats after fine-tuning our losses for each specific video }
    \end{longtable}
  
\end{tiny}



\end{document}